\documentclass[fleqn,10pt,twocolumn]{wlscirep}

\usepackage[utf8]{inputenc}
\usepackage[T1]{fontenc}
\usepackage{multirow}
\usepackage{dutchcal}
\usepackage{svg}%
\usepackage{threeparttable}%
\usepackage{marvosym} 
\title{Cooperative Network Learning for Large-Scale and Decentralized Graphs}
\author[1,*]{Qiang Wu}
\author[1,2,*]{Yiming Huang}
\author[1,2]{Yujie Zeng}
\author[1]{Yijie Teng}
\author[1,2]{Fang Zhou}
\author[3,1,2 \Letter]{Linyuan Lü}
\affil[1]{Institute of Fundamental and Frontier Sciences, University of Electronic Science and Technology of China, Chengdu 611731, China}
\affil[2]{Yangtze Delta Region Institute (Huzhou), University of Electronic Science and Technology of China, Huzhou 313001, China}
\affil[3]{School of Cyber Science and Technology, University of Science and Technology of China, Hefei 230026, China}
\affil[*]{These authors contributed equally to this work}
\affil[\Letter]{email: Linyuan.lv@ustc.edu.cn}

\begin{abstract}

Graph research, the systematic study of interconnected data points represented as graphs, plays a vital role in capturing intricate relationships within networked systems. 
However, in the real world, as graphs scale up, concerns about data security among different data-owning agencies arise, hindering information sharing and, ultimately, the utilization of graph data. 
Therefore, establishing a mutual trust mechanism among graph agencies is crucial for unlocking the full potential of graphs.
Here, we introduce a Cooperative Network Learning (CNL) framework to ensure secure graph computing for various graph tasks.
Essentially, this CNL framework unifies the local and global perspectives of GNN computing with distributed data for an agency by virtually connecting all participating agencies as a global graph without a fixed central coordinator.
Inter-agency computing is protected by various technologies inherent in our framework, including homomorphic encryption and secure transmission.
Moreover, each agency has a fair right to design or employ various graph learning models from its local or global perspective.
Thus, CNL can collaboratively train GNN models based on decentralized graphs inferred from local and global graphs.
Experiments on contagion dynamics prediction and 
traditional graph tasks (i.e., node classification and link prediction) demonstrate that our CNL architecture
outperforms state-of-the-art GNNs developed at individual sites, revealing that CNL can provide a reliable, fair, secure, privacy-preserving, and global perspective to build effective and personalized models for network applications.
We hope this framework will address privacy concerns in graph-related research and integrate decentralized graph data structures to benefit the network research community in cooperation and innovation.

\end{abstract}

\begin{document}

\flushbottom
\maketitle

\thispagestyle{empty}

\section*{Introduction}

Graph research, the systematic study of interconnected data points represented as graphs, aims to uncover hidden patterns and relationships within complex systems, holding the potential to drive advancements across numerous fields, encompassing healthcare and disease control\cite{health2020global-Science}, social connectivity\cite{social-nature}, environmental conservation\cite{martinez2006complex-envrionment-nature}, urban planning\cite{Urban-Science}, and economic growth\cite{hidalgo2011network-economic-nature}, among others.
However, traditional graph research models like Breadth-First Search and Depth-First Search, often use predefined rules and algorithms for tasks and do not typically learn from the data, leading to several limitations, such as the inability to capture complex relationships, scalability challenges, restricted flexibility, and sensitivity to noisy and incomplete data\cite{scarselli2009graph-limit,wu2020comprehensive-limit}. These constraints may hinder their ability to effectively handle the inherent complexities in graph-structured data.

Graph neural networks (GNNs), a class of machine learning models, provide a versatile and powerful tool for understanding graph data.
By aggregating information from neighboring nodes through multiple layers of computation, GNNs are capable of modeling intricate relationships within graphs\cite{HiGCN}, even in the presence of noisy and incomplete data.
Additionally, certain GNN variants are scalable, enabling efficient computation on large graphs by distributing the workload across multiple processing units. 
However, in large-scale graph tasks that involve multiple data agencies, privacy concerns\cite{Kaissis:2020, Price:2019} stemming from regulatory restrictions as well as competition in research and commercial domains may arise and prevent agencies from information exchange,  thus limiting the effectiveness of GNNs. 

Federated graph learning\cite{Zheng:2021, He:Nips2022, Wu:FL2022, Xia:FL2022} addresses privacy concerns associated with GNNs. 
It is built upon federated learning\cite{Kone:FL2016, yang:fl2019}, a novel machine learning approach that allows clients to collaboratively train a shared model under the guidance of a central server while keeping their data on their respective devices. 
However, this approach requires that each distributed agency upload its model's parameters or gradients to a central coordinator, which necessitates that these distributed agencies employ the same GNN model structure dictated by the coordinator. 
This centralized control bestows significant authority upon the coordinator, potentially reducing other agencies' enthusiasm for participation in research and development.
Moreover, exchanging gradients can inadvertently reveal private information, thereby increasing the risk of data leakage\cite{Zhu-gradient-2019}.

Swarm learning \cite{Warnat:swarm2021}, another decentralized machine learning approach, uniquely combines edge computing with blockchain-based peer-to-peer networking, enabling coordination without a central coordinator.
Here, users can securely share parameters via the swarm network and independently construct models on private data at individual sites, thereby providing robust security measures to ensure data privacy and confidentiality.
Yet, swarm learning, specially designed for decentralized and confidential clinical machine learning, also has its shortcomings. For example, it is unsuitable for non-I.I.D (independently and identically distributed) data like complex graph structures, and all nodes must utilize the same machine learning model as federated graph learning.
In other words, neither federated graph learning nor swarm learning can create flexible and diverse models for each participating agency with varying network structures. How to fairly coordinate network structures and data across different agencies while safeguarding data privacy remains a question. 
The proposed solution is anticipated to include several key features:
(1) Integrates graph data from any owner in the field of graph applications;
(2) Unifies global and local perspectives for diverse graph applications;
(3) Encourages diversity in graph algorithms for a wide range of applications;
(4) Retains graph data and algorithms locally with their respective owners;
(5) Ensures reliable, fair, secure, and privacy-preserving models for graph applications without violating privacy legislation;
(6) Eliminates the need for a central parameter coordinator, promoting equal rights among all participants;
(7) Provides a high level of security for both data and models using proven security technologies.

Here, we introduce Cooperative Network Learning (CNL), which unifies the formulation of graph models with distributed graph data from an integrated (local and global) perspective. This approach facilitates joint learning by virtually connecting all participating agencies as a global graph without a fixed central coordinator. 
Moreover, CNL can be generalized to address various graph problems by providing a secure graph computing framework based on homomorphic encryption\cite{Homomorphic:2009, homomorphic:2020}, thereby surpassing federated graph learning and swarm learning.
As a result, CNL enables each participating agency to fairly study and deploy its own dynamic GNN model from global and local perspectives.

To validate the practicality of the CNL approach, we conduct experiments on various graph learning tasks, including contagion dynamics prediction, node classification, and link prediction. 
Numerical experiments demonstrate that our CNL architecture offers a reliable, fair, secure, privacy-preserving, and global perspective for constructing accurate and practical models for graph applications.
We anticipate that this framework will overcome existing challenges and facilitate the integration of decentralized graph data structures, ultimately benefiting the network research community by fostering cooperation and innovation.

\begin{figure*}[h]
\centering
\includegraphics[width=\linewidth]{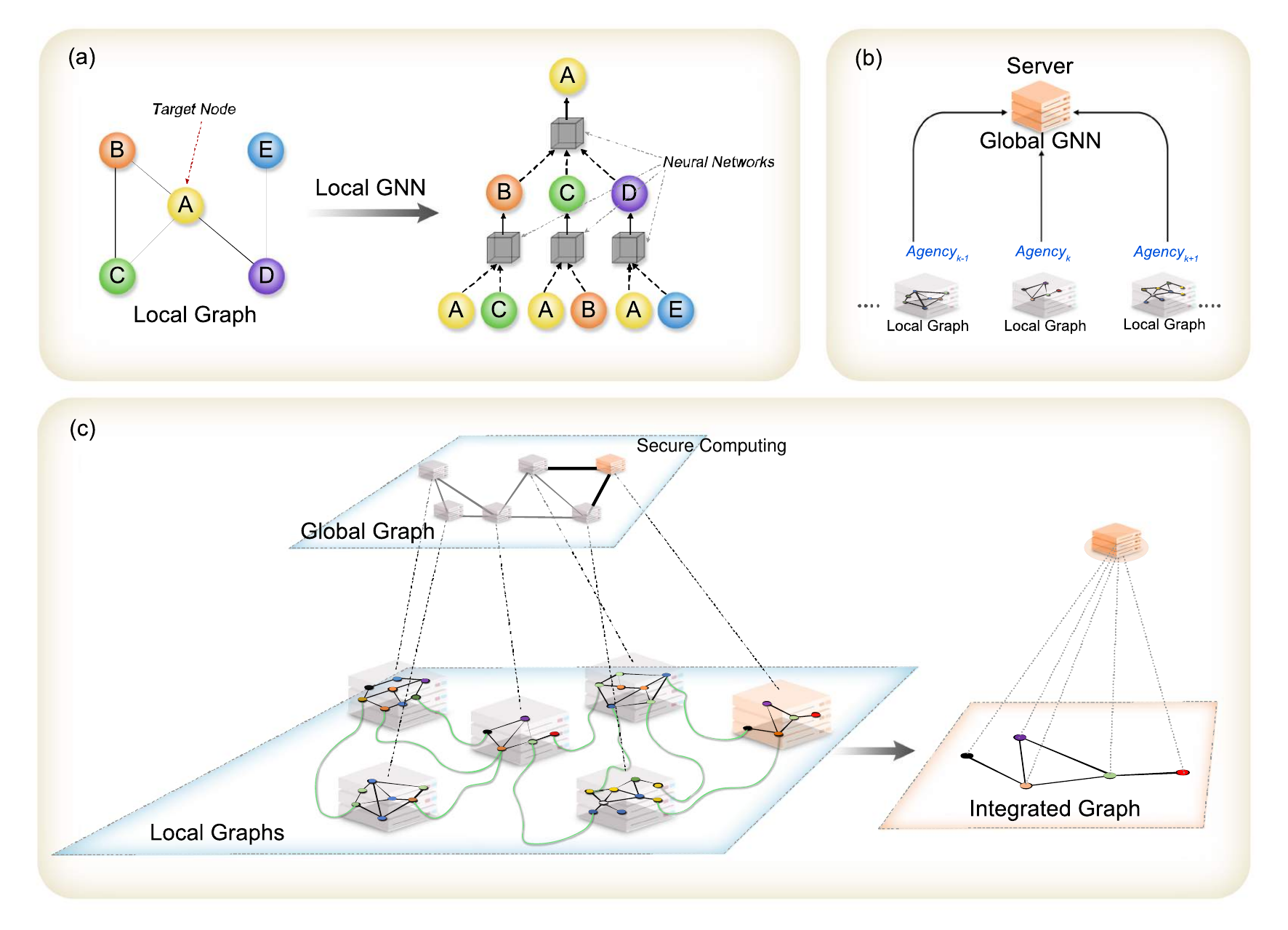}
\caption{
{\bf Cooperative Network Learning framework.} 
(\textbf{a}) Graph neural network with local graph data. 
(\textbf{b}) Federated graph learning with a central parameter server. It performs a shared graph neural network model on local and global graphs. 
(\textbf{c}) Cooperative Network Learning that consists of three perspectives: local, global, and integrated. 
Local GNN operates on a Local Graph, producing embeddings for nodes within the agency. Subsequently, Global GNN works on a Global Graph, generating embeddings for each agency. 
Integrated GNN operates on an Integrated Graph, capturing information outside the agency, resulting in more accurate embeddings. 
}
\label{fig:architecture}
\end{figure*}

\section*{Results}
\label{section:result}

\subsection*{An overview of the CNL framework }

In numerous instances, accessing and leveraging all data can be challenging due to privacy protection or commercial competition constraints.  
Typically, access is restricted to information intrinsic to a specific agency, i.e., limiting data acquisition within the local graph. 
Models instantiated under such constraints are designated as local GNNs (Fig. \ref{fig:architecture}(\textbf{a})).

As illustrated in Fig. \ref{fig:architecture}(\textbf{c}), our Cooperative Network Learning (CNL) framework combines secure computing and GNNs to facilitate privacy-preserving and equitable collaboration.
Specifically, first, local GNNs process the data in the local graph (owned by an individual agency) in a conventional GNN approach, and subsequently employ a graph pooling operation\cite{GCN2017} to derive the embedding for the agency associated with the local graph. 
Next, global GNNs process graph-level information from participating agencies within the global graph, whose node denotes an entire agency and whose edge reflects the relation between agencies. 
Finally, an integrated GNN model is generated among local and global graphs by connecting all local nodes in an agency with a virtual global node.
In the training of global and integrated GNNs, we employ multiple secure computing approaches including homomorphic encryption to ensure agencies update their graph embeddings without disclosing raw graph data (see "Methods" for details).

In general, local GNNs capture micro-level relationships between nodes within each agency, while global GNNs reveal macro-level relationships among agencies. 
By integrating both the macro and micro information on an agency's private server, the integrated GNN model helps preserve data privacy and ensure equal participation while accommodating various GNN algorithms.


\subsection*{Performance evaluation}
Depending on the scope of graph data GNNs operate on, there are two common types of models: local GNNs and centralized GNNs.
Local GNNs exclusively utilize data within their respective agencies, while centralized GNNs can access data of all agencies during their training procedures. 
Note that centralized GNNs represent an ideal scenario, as accessing and leveraging all data can be challenging due to privacy protection or commercial competition constraints.
It is also worth mentioning that the centralized GNNs are not equivalent to the global GNNs.
Our integrated GNNs incorporate local and global GNNs, utilizing not only local embedding learning within each agency's local graph but also the agency's embedding based on the global graph.
This integration contributes to improved performance compared to local GNNs.

%
To examine the performance of our framework, we have three evaluating angles for these models: 
single-node performance, which measures the performance of a single node; 
single-agency performance, which assesses the overall performance of all nodes within a single agency; 
multi-agency collective performance, which ideally considers the performance of all nodes across all agencies. 
Note that when conducting a comparative analysis, the performance of different models is assessed using the same angle to ensure fairness in the evaluation.


\subsection*{CNL on contagion dynamics}
Contagion dynamics prediction involves forecasting the spread and impact of infectious diseases, information, or trends within a network using mathematical models.
To assess the efficacy of the CNL framework, we employ various GNN models within this framework to predict the spreading impact of a disease on both synthetic and empirical networks (see Supplementary Notes $5$ and $6$ for details). 

Specifically, we elaborately design customized models within the CNL framework, which encompass CNNs\cite{Krizhevsky-2012}, i.e., the temporal information processing module, and GNN, i.e., the spatial information processing module. (see "Methods" for details about these customized models).
Moreover, we employ root mean square error (RMSE) and Pearson correlation coefficient (PCC) as two evaluation metrics, with improved prediction performance indicated by a decrease in RMSE and an increase in PCC. 

\begin{figure}[!h]
\centering
\includegraphics[width=\linewidth]{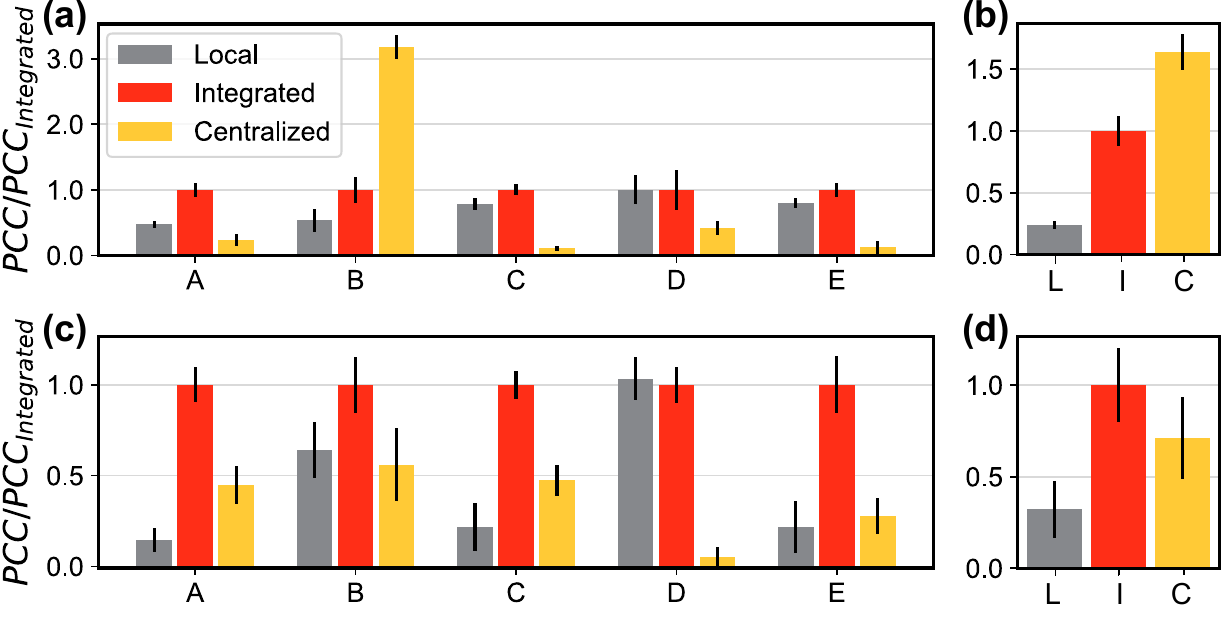}
\caption{{\bf Comparison of local, integrated and centralized models on contagion dynamics prediction.} 
The upper and lower panels represent performance on generalized ER and BA networks, respectively.
(\textbf{a}) and (\textbf{c}) illustrate the single-agency performance, where A-E represent five separate agencies, while (\textbf{b}) and (\textbf{d}) represent the multi-agency collective performance. 
For each subplot, the $y$-axis denotes the PCC value, and each value is scaled to one of the integrated model's PCC value for the same dataset. }
\label{fig:synthetic_contagion}
\end{figure}

In synthetic networks, we utilize Erd{\H{o}}s–R{\'e}nyi (ER) \cite{barabasi2013network} and Barab{\'a}si–Albert (BA)\cite{BA1999} graphs, where nodes are partitioned into five agencies with the spectral clustering algorithm \cite{ng2002spectral}. 
Besides, we employ the Susceptible-Infected-Recovered (SIR) model\cite{barabasi2013network} and Susceptible-Infected-Recovered (SIS) model\cite{morone2015influence} to simulate the epidemic progression.
The results are visualized in Fig. \ref{fig:synthetic_contagion}.

Integrated models typically outperform local models, largely due to their incorporation of agency embeddings in addition to local information. Intriguingly, despite their access to a more extensive dataset, centralized models occasionally fall short in performance compared to both local and integrated models for certain tasks.

Generally, the integrated models achieve a better performance than local models, largely due to their incorporation of agency embeddings in addition to local information.
We also note that the centralized models, while acquiring more data, do not perform as well as local and integrated models in some tasks. 
Specifically, as shown in Fig. \ref{fig:synthetic_contagion} ({\bf a}) and ({\bf b}), in the ER network, the multi-agency collective performance of the centralized model is about 1.5 times greater than that of the integrated model. 
This is mainly because the performance of the centralized model in agency B is pretty high.
In addition, in the remaining four agencies A, C, D, and E, integrated models perform 1-2 times better than local models, and 2-9 times better than centralized models.
As shown in Fig. \ref{fig:synthetic_contagion} ({\bf c}) and ({\bf d}), in the BA network, the integrated model achieves the best results in 4 out of the 5 agencies, and performs up to 6.8 times better than the local model in the best situation. 
In addition, the integrated model outperforms the centralized model in all five agencies, which reflects that our model can get better performance in all agencies even with incomplete data.
Additional experiments with synthetic contagion data are deferred to Supplementary Note $8$ in Tabs. S3 and S4.

\begin{figure}[ht]
\centering
\includegraphics[width=1\linewidth]{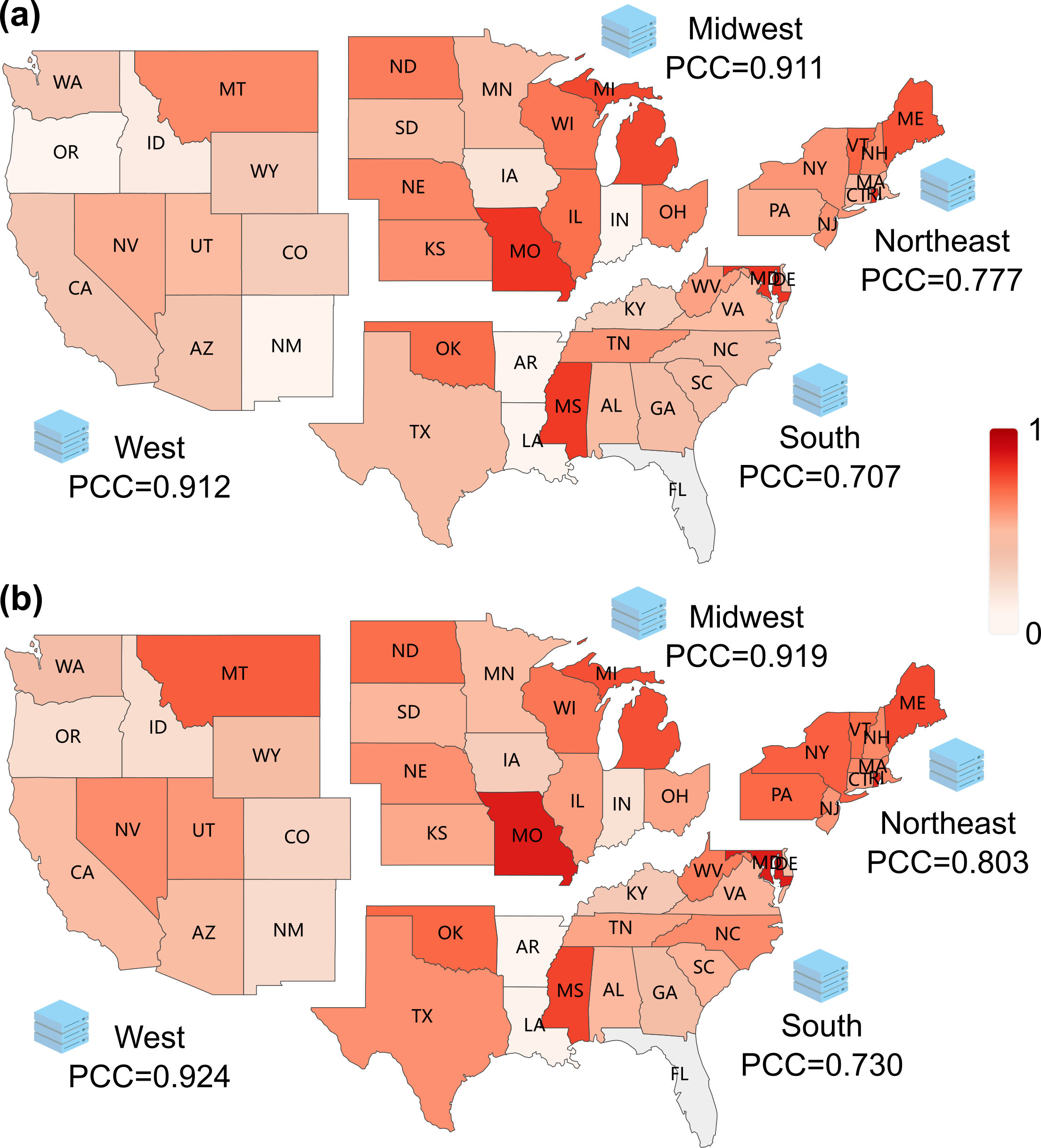}
\caption{\textbf{Comparison of local and integrated models on contagion dynamics prediction.} 
Panels (\textbf{a}) and (\textbf{b}) show the performance of local and integrated models in predicting contagion dynamics on the US state dataset, respectively. 
The performance is measured using the PCC metric at both the single-node level (each state represents a node) and the single-agency level (encompassing four regions: northeast, midwest, south, and west).
Node performance is color-coded,  with darker red indicating better performance, and agency performance is represented by numerical PCC values, with higher values indicating better performance.}
\label{fig: state360_map}
\end{figure}

As to the contagion dynamics prediction on empirical networks, we conduct experiments on a US influenza dataset. Here, states are partitioned into four agencies based on their geographical location, and Fig. \ref{fig: state360_map} visualizes the prediction results.
We can draw that the integrated model based on our CNL framework outperforms all local models in single-agency performance, although the magnitude of the performance gain for the nodes within the agencies may vary.

For a more extensive comparison, we also evaluate our model using three other epidemic-related datasets: US-Region, Spain, and Twitter (see Supplementary Notes $6$ for details).
In addition to geographic location, we also employ the spectral clustering algorithm \cite{ng2002spectral} for agency partitioning.
Furthermore, except for the GAT module, which serves as the spatial information processing component in our previous experiments on contagion dynamics, we also integrate other modules into the CNL framework: GAT\cite{Petar:2018gat} and GraphSAGE\cite{Hamilton:2017graphsage} (see Supplementary Notes $7$ for details).
The fact that CNL can adapt to various agency partitioning methods and different information processing modules highlights its remarkable versatility.

\begin{table*}[ht]
\centering
\caption{{\bf Comparison of customized CNL models and baseline methods on contagion dynamics prediction.}}
\label{tab: empirical_rel}
\begin{threeparttable}
\begin{tabular}{lllllllll}
\toprule
\multirow{2}{*}{Method} 
& \multicolumn{2}{c}{US-States} 
& \multicolumn{2}{c}{US-Regions} 
& \multicolumn{2}{c}{Spain} 
& \multicolumn{2}{c}{Twitter} \\
\cmidrule(r){2-3} \cmidrule(r){4-5} \cmidrule(r){6-7} \cmidrule(r){8-9}
                            & RMSE  & PCC       & RMSE  & PCC       & RMSE  & PCC   & RMSE  & PCC \\ \midrule
AR $\clubsuit$              & 306   & 0.773     & 1330  & 0.612     & 218   & 0.045 &1380   &0.955\\
LSTnet $\blacktriangle$     & 292   & 0.760     & 1157  & 0.609     & 215   & 0.163 &1315   &0.959 \\
ST-GCN $\blacktriangle$     & 289   & 0.769     & 1290  & 0.644     & 176   & 0.278 &1331   &0.959 \\
EpiGNN $\blacktriangle$     & 220   & 0.865     & 984   & 0.749     & 175   & 0.308 &1302   &0.960\\
ColaGNN $\blacktriangle$    & 214   & 0.822     & 1134  & 0.717     & 167   & 0.397 &1329   &0.959\\
CNNRNN-Res $\blacktriangle$ & 260   & 0.820     & 1233  & 0.552     & 182   & 0.049 &1321   &0.958\\
CNL-SAGE $\clubsuit$        & 235   & 0.851     & 983   & 0.724     & 231   & 0.149 &1299   &0.959\\
CNL-SAGE $\bigstar$         & {\bf 234}   & {\bf 0.852}     & {\bf 936}   & {\bf 0.765}     & {\bf 171}   & {\bf 0.401} &{\bf 1284}   &{\bf 0.961}\\
CNL-SAGE $\blacktriangle$   & 218   & 0.872     & 1098  & 0.701     & 172   & 0.319 &1330   &0.958\\
CNL-GAT $\clubsuit$         & 266   & 0.805     & 915   & 0.768     & 192   & 0.236 &1455   &0.865\\
CNL-GAT $\bigstar$          & {\bf 242}   & {\bf 0.849}     & {\bf 896}   & {\bf 0.779}     & {\bf 188}   & {\bf 0.275} &{\bf 1398}   &{\bf 0.876}\\
CNL-GAT $\blacktriangle$    & 224   & 0.865     & 875   & 0.789     & 172   & 0.386 &1341   &0.959\\ \bottomrule
\end{tabular}

\begin{tablenotes}
\item Multi-agency collective performance of local ($\clubsuit$), integrated ($\bigstar$), and centralized ($\blacktriangle$) models.
In the CNL framework, compared to local GNN models, the integrated GNN models with improved performance are highlighted in bold.
\end{tablenotes}
\end{threeparttable}

\end{table*}

Table \ref{tab: empirical_rel} summarizes the results of various methods in terms of RMSE and PCC. The substantial variation in RMSE across datasets can be attributed to differences in dataset size and variance.
The results suggest that customized models powered by the CNL architecture outperform most centralized models or approach the best performance of centralized models. Moreover, integrated models generally outperform local models in CNL.

\subsection*{CNL on traditional graph tasks}

To determine whether CNL is a general framework applicable to various graph learning tasks beyond contagion tasks, we further apply it to two standard graph learning tasks: node classification and link prediction.

Node classification is an essential task in many network science applications, including predicting protein functions from interaction networks, identifying spammers in online social networks, and detecting fraudulent users in financial transaction networks\cite{Senevirathne:IEEEgnn2020}. Accurate node classification can significantly improve the performance of various real-world systems and applications\cite{GCN2017} (see Supplementary Note $5$ for detailed descriptions).

\begin{figure}[!ht]
\includegraphics[width=\linewidth]{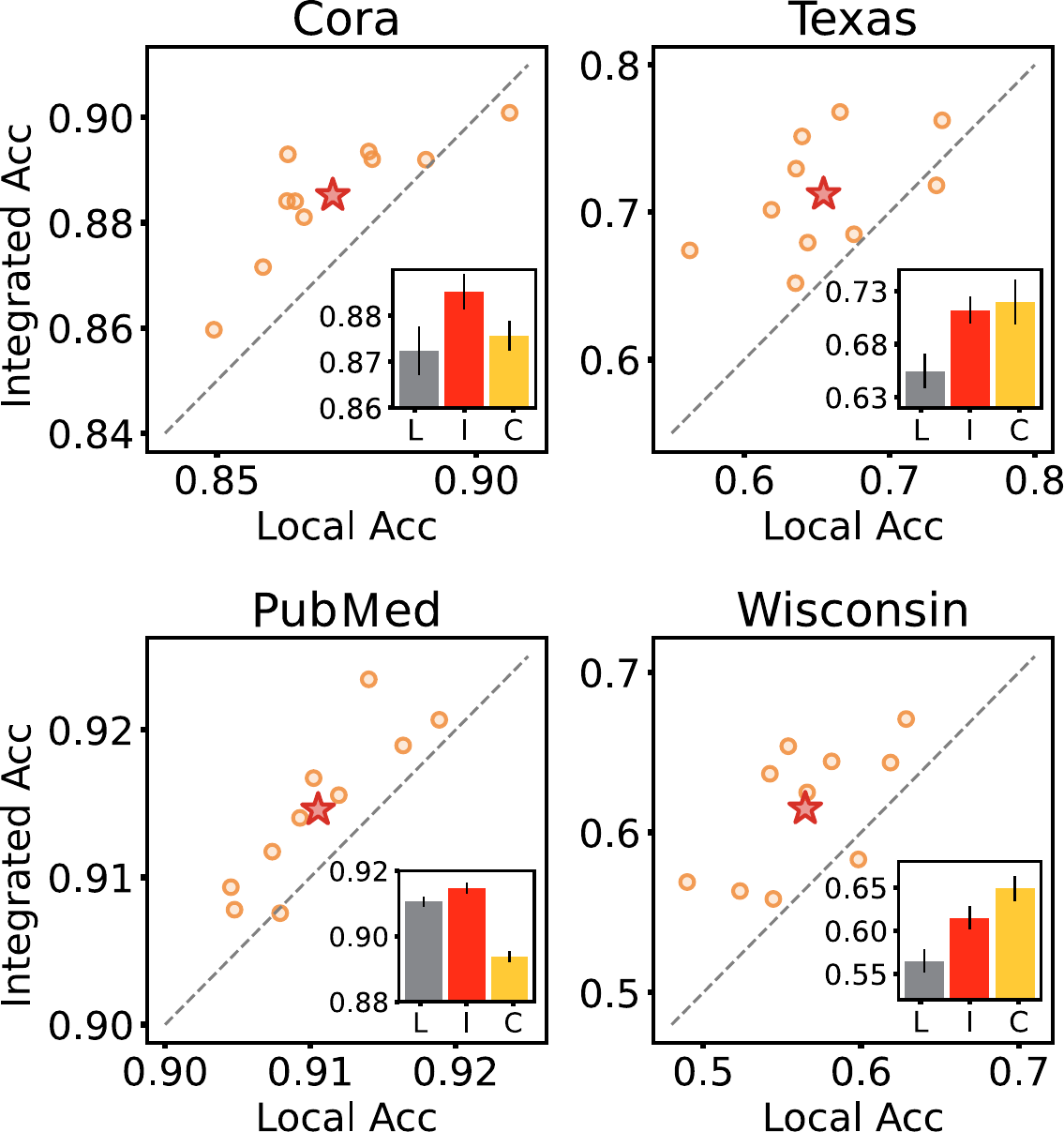}
\caption{{\bf Comparison of local, integrated, and centralized models on node classification.}  The four subplots represent models' multi-collective agency performance on four datasets, i.e., the Cora, Texas, PubMed, and Wisconsin, respectively.  
For each subplot, the x-axis denotes the accuracy of local models, and the y-axis represents the accuracy of integrated models. The ten dots indicate 10 repeated experiments, with their mean value visualized by an asterisk. The bar chart in the bottom right shows the performance of the local, integrated, and centralized models together. 
Cora and PubMed in the left panel are homogeneous datasets, while Texas and Wisconsin in the right panel are heterogeneous datasets. }
\label{fig:node_classify}
\end{figure}

To better validate CNL's effectiveness, we carry out experiments on four different datasets: two homogeneous graphs, i.e., Cora \cite{data:Cora} and PubMed \cite{data:PubMed}, and two heterogeneous graphs, i.e., Texas and Wisconsin \cite{data:Pei2020Geom-GCN} (see Supplementary Note $6$ for details). 
For each dataset, we use spectral clustering\cite{ng2002spectral} to partition the nodes in the adjacency matrix into three distinct groups, corresponding to agencies A, B, and C, respectively. 
For this task, we integrate the Graph Convolutional Network (GCN) model\cite{GCN2017}, a renowned GNN model, into the CNL framework as the spatial information processing module. 
We use the CNL framework to securely transfer the embedding vectors obtained from the GCN model to obtain the Integrated model. 
We use average classification accuracy (ACC) as the metric, and the results of the local, integrated, and centralized models are shown in Fig. \ref{fig:node_classify}.

The results show that the performance of the Integrated models generally surpasses that of the Local models.
Moreover, in the homogeneous graphs, the distinctions among the Local, Integrated, and Centralized models are nuanced. 
The augmentation observed in the Integrated model is relatively slight compared to the Local model. Intriguingly, the Centralized model doesn't surpass the Local counterpart. 
Conversely, when navigating heterogeneous graphs, the performance disparities among these three models become markedly evident. The Centralized model performs the best, followed by the Integrated model, which outperforms the Local model.
Through this extensive examination, we successfully validate the capabilities of the CNL framework, demonstrating its potential to enhance node classification tasks and its adaptability across diverse graph architectures.


In addition to node-level tasks, we consider the performance of CNL on link prediction, a classical edge-level task that focuses on the relationships between nodes. 
Link prediction techniques have been applied to problems ranging from suggesting new friends on social networks to identifying proteins that might bind to a particular drug\cite{liben2007link}. Accurate link prediction can uncover hidden relationships and patterns in data, leading to better decision-making and improved system performance\cite{liben2007link} (see Supplementary Note $5$ for detailed descriptions).

\begin{figure}[!ht]
\includegraphics[width=\linewidth]{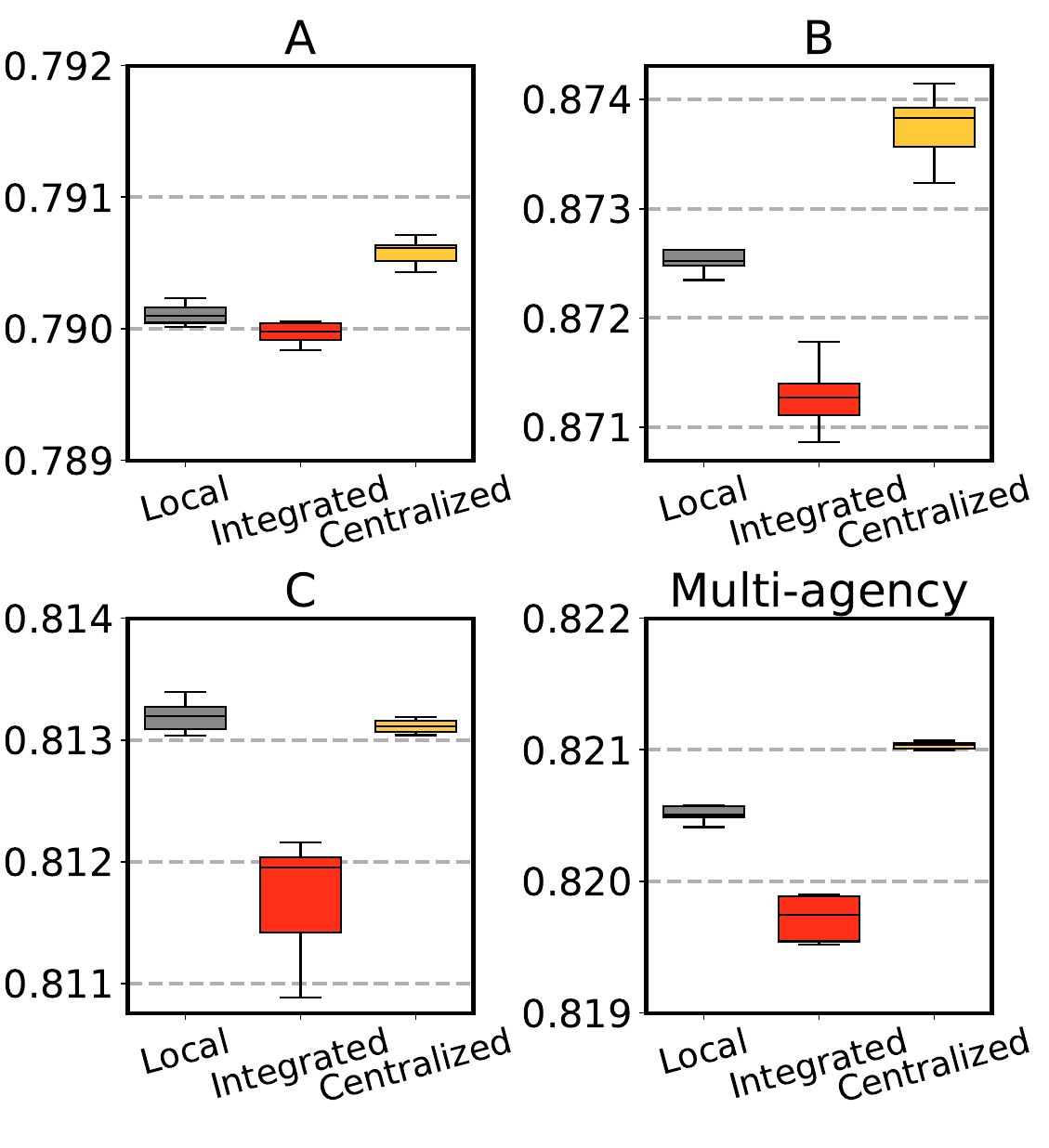}
\caption{\textbf{Comparison of local, integrated, and centralized models on link prediction.} 
The four subplots display three models' single-agency (A, B, C) performance and multi-agency collective performance on the Ciao dataset, respectively. 
Each experiment is repeated $10$ times. For each subplot, the y-axis represents the MAE values.
The box, with a median line inside, represents the interquartile range (IQR), while the whiskers extend from the box to show the minimum and maximum values.
 }
\label{fig:edge_classify}
\end{figure}

Here, we conduct experiments on a recommender system, specifically using the Ciao dataset\cite{tang2012mtrust}, where users rate products on a scale of 1-5 stars (see Supplementary Note $6$ for details). Our objective is to assess CNL's ability to infer users' ratings for products, essentially predicting edge weights.  
Following the partitioning method proposed in this work\cite{he2021fedgraphnn}, we divide the dataset into three parts based on product categorization, each representing an independent agency in the CNL. 
Similar to the node classification task, we also employ GCN as the spatial information processing module. 
Using the mean absolute error (MAE) as the evaluation metric, we evaluate the Local, Integrated and Centralized models on each of the three agencies and the entire graph. The results are illustrated in Fig. \ref{fig:edge_classify}.

Overall, the integrated model achieves the best performance on link weight prediction, as indicated by the minimum MAE values for all three agencies (see Supplementary Note 8 Table S9 for details), outperforming both the centralized and local models. 
Specifically, the integrated model consistently achieves the lowest median MAE across all three agencies and the entire graph. It also achieves the lowest average MAE in nearly all cases. 
%


The finding that the integrated model outperforms the local model is reasonable since the former utilizes neighbor agencies' information, which the latter lacks. However, the inferior performance of the centralized model can be attributed to the presence of an outlier. Notably, the Ciao dataset incorporates product categorization, and the adopted partitioning scheme tends to aggregate products of the same category into the same subgraph. Because users show diverse evaluation patterns for products across different categories\cite{gao2023impact}, the centralized model, relying on all available information, is vulnerable to interference. In contrast, the integrated model, leveraging extracted common features, enhances predictive performance. This outcome aligns with a previous study\cite{he2021fedgraphnn}.

\section*{Discussion}
In the rapidly developing field of network research, the Collaborative Network Learning (CNL) system stands out as a promising departure from the traditional centralized learning or parameter-sharing paradigms. Its decentralized graph learning system holds the potential to revolutionize our understanding and utilization of complex network structures and dynamics.

CNL's robustness against data security threats, and its resilience to non-contributing or potentially malicious participants, can be attributed to its utilization of homomorphic encryption\cite{Homomorphic:2009,Froelicher:2021,homomorphic:2020}.
This sophisticated approach fortifies the CNL network and its node embeddings, serving as a bulwark against potential attacks.  

Besides, unifying the formulation of graph models with distributed graph data, the CNL framework presents a novel approach to capturing various graph information. 
 Specifically, it virtually links all participating agencies into a global graph, seamlessly integrating the global and local perspectives of dynamic processes, which is achievable for any node within a participating agency.

Moreover, our CNL system is decentralized, with no single agency having the authority to approve or reject the participation of other agencies in the CNL framework. Without a central parameter server, each agency retains autonomy in designing or employing dynamic models on their local or global graphs. This levels the playing field, ensuring equal rights for all agencies and fostering a more cooperative and democratic network learning environment.

Another significant advancement of our CNL model is that it can securely collect and integrate graph data from multiple agencies without compromising data privacy, surpassing federated graph learning and swarm learning. Here, both the graph data and global encrypted embedding are processed within each agency's server, making the whole process easier, more efficient, and safer.

Numerical experiments have demonstrated that our CNL framework surpasses traditional GNN models, whether trained on empirical or synthetic graph datasets. 
In our experiments, a customized GNN model, trained on the CNL framework with a global perspective, always outperforms models trained on local graph data alone. Sometimes, it even performs better than centralized GNN models, as it may eliminate extraneous noise information that might otherwise interfere with the results.

To the best of our knowledge, CNL pioneers a distributed artificial intelligence security computing solution without relying on a center coordinator or parameter sharing.
Apart from its exceptional performance in contagion dynamics and traditional graph tasks, CNL has the potential to extend to more graph tasks to improve its generalizability. 
For instance, community detection and graph generation could be integrated into the model to provide a more comprehensive understanding of network structures. We can also apply CNL to subgraph pattern mining, helping facilitate the discovery of frequent, discriminative, and dynamic patterns in graph data. Moreover, the CNL framework could be deployed for complex network evolution prediction, which may significantly contribute to the anticipation and management of changes in network systems. These extensions will undoubtedly enhance the universality of the CNL model, promising significant advancements in complex network research.

Additionally, the CNL model exhibits certain limitations in its explanatory capability. First, it fails to provide a clear rationale for the varying performance observed at both the node and agency levels, where some nodes or agencies experience improvement while others either stagnate or deteriorate. Furthermore, the model needs to be more precise in elucidating why it outperforms centralized GNN in specific tasks, lacking an analytical exploration into its behavior across different graph attributes.
We anticipate that more research will build upon this foundation to delve deeper into this vital yet challenging aspect of network science.

\section*{Methods}
\subsection*{ Computing architecture of CNL}

Our CNL framework employs GNN models that compute from three perspectives  (Fig. \ref{fig:architecture}({\bf c})): a local perspective that uses a local GNN for each agency, a global perspective that leverages a global GNN for all agencies, and an integrated perspective that combines both global and local perspectives under privacy protection.

\begin{figure*}[ht]
\centering
\includegraphics[width=\linewidth]{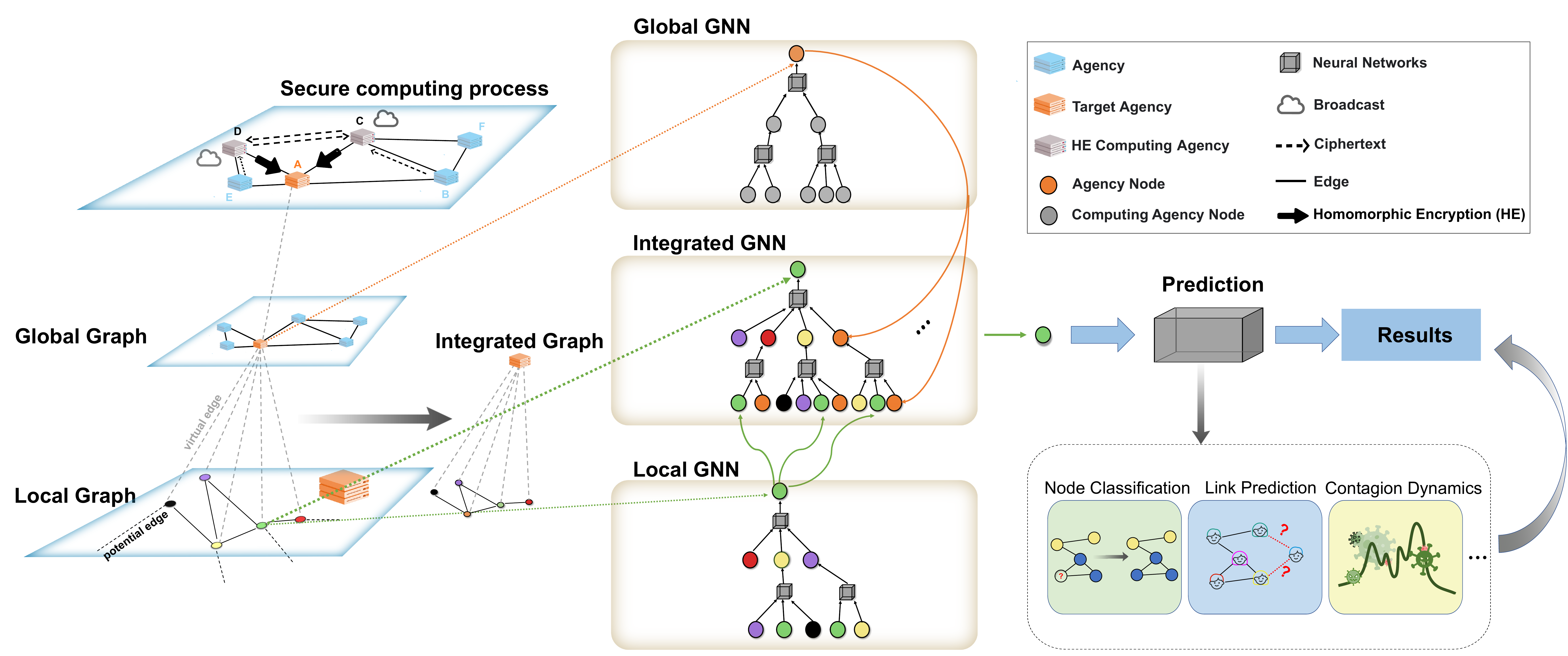}
\caption{{\bf The running process of local, integrated, and global GNNs.} 
The global GNN is performed on the global graph with homomorphic encryption technology to generate global graph embeddings.
The integrated graph consists of the participating agency (graph level) as the center node, virtually connecting all nodes (node level) in the local graph.
It locally stores the local graph and integrated graph, where local GNNs and integrated GNNs are performing on its own server.
The local node embedding is updated by its neighbors and the virtual center node (graph embedding). 
Based on the local node embedding, the combined macro and micro results are predicted by an integrated GNN model designed by the agency.
}
\label{fig:integrated-GNN}
\end{figure*}

In the local graph perspective, nodes and edges are specifically owned by an individual agency; the local GNN aims to find the micro-level information between nodes within each agency and generates embedding for each node. On the other hand, in the global graph perspective, each participant agency is regarded as a node; the global GNN intends to capture the macro-level interaction between these agencies. Thus, the agencies are connected by distributed computing\cite{Farashahi:2021} with homomorphic encryption\cite{Homomorphic:2009,Froelicher:2021,homomorphic:2020} for global topology.

In the integrated graph perspective, a global agency node virtually connects all nodes in its local graph, resulting in an integrated graph (Fig. \ref{fig:architecture}({\bf c})).
The integrated GNN model is designed to capture both the micro and macro information of an agency. Specifically, the local node embeddings it uses have already incorporated the global characteristics of this agency.
Moreover, the integrated GNN model is computed within each agency's private server, operating without the need for a central coordinator. 
This allows the participating agency to maintain the privacy of their raw graph data and ensures equal rights for enrolling and computing.
Additionally, various GNN algorithms could be developed for the integrated graph to obtain the final results considering macro and micro information.

For more details about the three-level computing process of GNNS, please refer to Supplementary Note $1$.

\subsection*{Data security and privacy protection}

To guarantee comprehensive data security and privacy, our CNL framework primarily adopts a homomorphic encryption algorithm to integrate the embedding of neighboring agencies in the computing process, as shown in Fig. \ref{fig:integrated-GNN}. Moreover, the reduction of information during transmission, coupled with the decentralized service for agencies, also helps ensure data security. 

Homomorphic cryptography\cite{Homomorphic:2009, Froelicher:2021, homomorphic:2020} enables users to perform mathematical operations on encrypted data, producing the same result as if the operations were performed on the original data. This is achieved by transforming plaintext data into ciphertext data, which can be manipulated without revealing the original data, and then decrypting the ciphertext data to obtain the final result. 
This means that operations such as retrieval and comparison can be performed on encrypted data without the need to decrypt it, thus maintaining the privacy of the data throughout the entire process.

Specifically, CNL uses the Paillier scheme\cite{paillier1999public} for homomorphic encryption, ensuring compliance with the standard security definition for encryption schemes: semantic security, also known as indistinguishability under chosen-plaintext attack (IND-CPA)\cite{Abdalla-2014}. 
%
Mathematically, homomorphic cryptography can be expressed as follows: Let $E(\cdot)$ and $D(\cdot)$ denote  the encryption and decryption functions, respectively, and the encryption operations satisfy
\begin{equation}
\label{equ:1}
     E(x) + E(y) = E(x+y).
\end{equation}
Then this pair of processes meet the homomorphic addition operation\cite{paillier1999public}, namely
$ x + y = D\left(E(x)+E(y)\right)$.

To illustrate, suppose we have the embeddings of two agencies, denoted as $x$ and $y$, and need to calculate the embedding of another agency, which is the sum of the embeddings of the first two, i.e., $(x+y)$.
Using homomorphic encryption, we can perform this computation on the encrypted embeddings of $x$ and $y$, without revealing their actual values.
Since $x + y = D\left(E(x)+E(y)\right)$, only the obtained result $(x+y)$ is available in the global GNN, without revealing the individual embedded values of $x$ and $y$.
Consequently, this approach ensures data security properties in our CNL framework.

Besides, the information will be lost when transmitted through the CNL framework, which also contributes to data privacy.
In the CNL framework, data transmission between agencies relied on graph embedding, a method for representing graph data in a lower-dimensional space while preserving its structure. According to Shannon's information theory\cite{shannon-1948}, the transmission process increases the uncertainty of the embedding compared to the original data, substantially reduces information content, and makes it difficult to infer the original information from the embedding, thereby ensuring data security. 
Moreover, CNL randomly selects several agencies as HE computing agencies for homomorphic ciphertext summation, and neighboring agencies can randomly choose some of them to send their own homomorphic encrypted embedding. 
Note that the granularity of the encryption can be adjusted by setting different numbers of HE computing agencies. 
This operation makes it extremely challenging to infer or trace specific neighbor embeddings and further enhances privacy protection.
In addition, in the CNL framework, there is no central node or central service that stores extensive information. Instead, neighbors directly interact with each other, and data owners always hold their original data, with encryption applied when processing is necessary. This initiative ensures fair access and usage while safeguarding data.

In Supplementary Note 3, we present the detailed working flow of computing security of the CNL framework under homomorphic encryption.

\subsection*{Cooperative network node service}

The Cooperative Network Node Service (CNNS) is the foundation of the CNL framework and the main program component as well.
It can execute various tasks, including communication, configuration parsing, and embedding model training, with the help of advanced homomorphic encryption modules.
CNNS deploys a robust client/server model via remote procedure call over the TCP/HTTP network protocol.
Specifically, the client dispatches a call message with the task parameters to the server, which then executes tasks based on the provided parameters and selectively returns a response. 

Within the CNL framework, CNNS is perpetually active, performing dual roles as clients and servers. 
The parameters, encrypted by  Rivest–Shamir–Adleman (RSA) public key encryption, are first transmitted by the CNNS client-side and are subsequently decrypted and executed by the server-side CNNS upon receipt.
Besides, a service-configuration separation principle governs the CNL framework. 
 CNNS relies on real-world inputs and necessitates specific configurations during runtime. The CNL bifurcates into a user area containing configuration and dataset directories and a firmware area allocated for the CNNS service. Thus, CNNS operates as an autonomous Docker container program service, detached from configuration and data. For more descriptions about CNNS, please refer to Supplementary Note $4$.

\subsection*{Customized modeling in CNL}
A distinctive trait of the CNL framework lies in its inherent flexibility, which enables each participating agency to design or employ a unique GNN model.
As an example, when dealing with contagion prediction, we develop a customized model specifically for this task, which comprises a temporal information processing module, namely Convolutional Neural Networks (CNNs), a spatial information processing module, namely Graph Neural Networks (GNNs), and a decoder.
CNNs\cite{Krizhevsky-2012} are a type of neural network that excels in feature representation and efficient parallel computation for sequential data due to their utilization of adaptable filters, which enables them to discern the underlying patterns in the data. 
To tailor the model for our contagion prediction task, we deploy multiscale convolutions as feature extractors to capture complex temporal patterns simultaneously.

\begin{figure}[!t]
\includegraphics[width=1\linewidth,height=5cm]{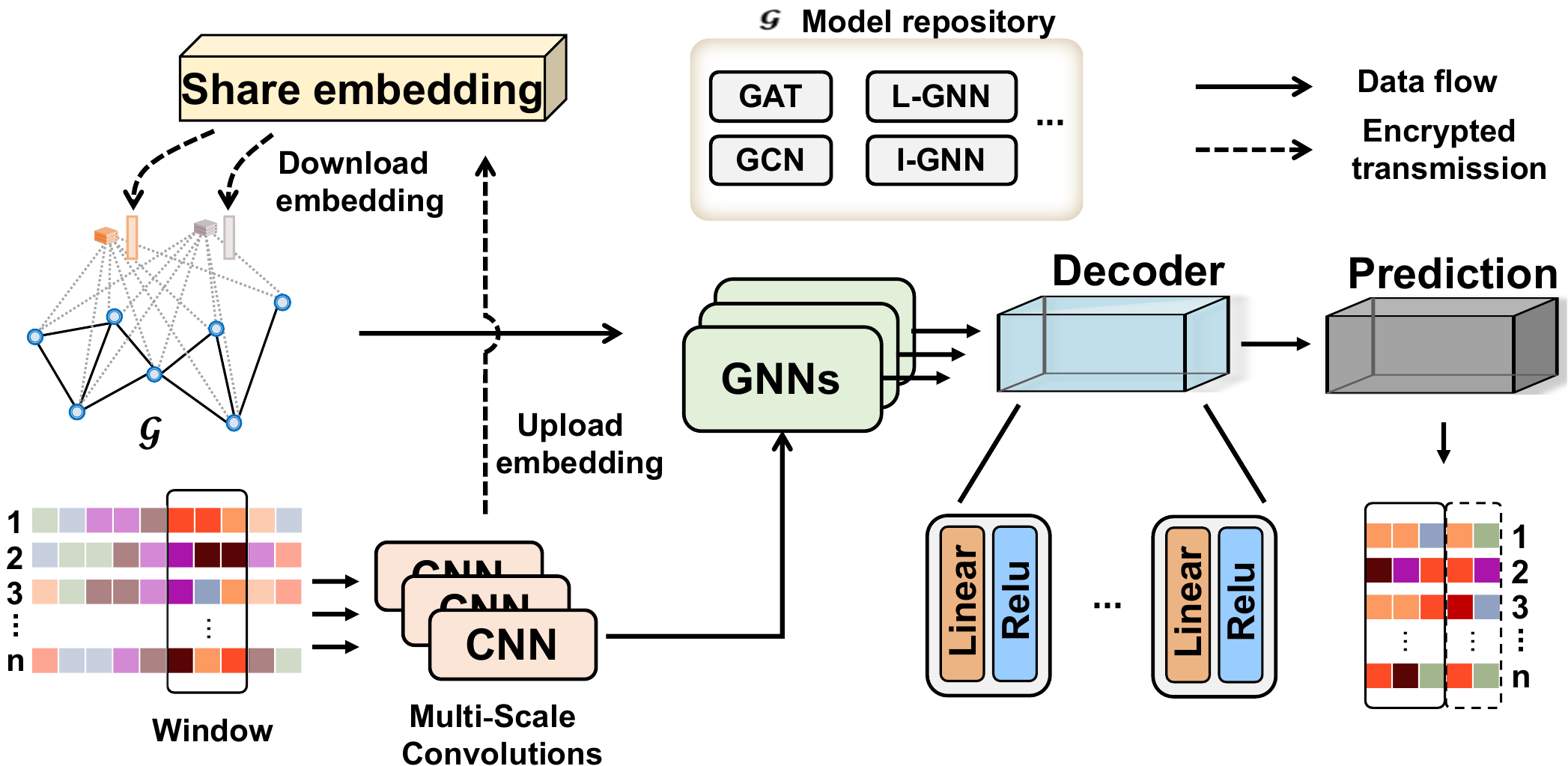}
\caption{\textbf{The customized model for the contagion dynamics prediction task.} 
We employ multiscale convolutions with different filter sizes and dilated factors as feature extractors, followed by GNNs to fuse information from different locations. The final prediction is then produced upon the application of a decoder.}
\label{fig:personalized_model}
\end{figure}

Specifically, as depicted in Fig. \ref{fig:personalized_model},  we use $\left[x_{t-T+1},\cdots,x_t\right] \in \mathbb{R}^{N\times T}$ for a specific look-back window $T$ to predict $x_{t+h}$,  where $x_\tau\in \mathbb{R}^N$ represents the observed case values of all nodes at time $\tau$.
The temporal features acquired by the CNNs are securely transmitted to neighboring agencies through the robust framework of CNL, and the distinct agency embeddings are treated as distinct virtual nodes in the customized model.
Concurrently, each agency can leverage the embedding information shared by other agencies. 
In the subsequent stage, we employ diverse GNN models to examine the dynamic spread of contagion across various regions (including the virtual nodes).
The participating agencies are programmed to fine-tune their integrated GNN model when observing a reduction in the validation set loss. The training procedure stops when the performance metrics across all agencies cease to exhibit improvement. See the results of the customized model in the previous \textbf{Section Results}.
Please consult Supplementary Note $7$ for information regarding the customized models utilized in node classification and link prediction tasks.

\clearpage
\section*{Data availability}
The data corpus used in this research consists entirely of publicly accessible datasets. All experiments and analyses were conducted in compliance with their respective original licenses. We have aggregated these public datasets alongside our self-generated ones at this link: \href{https://drive.google.com/drive/folders/1wTwpj3XzqzySTIDDew9LVkb7_tgiwypU?usp=sharing}{Google Drive Link}.

\section*{Code availability}
The code for this research can be accessed via the link: \href{https://github.com/CooperativeNetworkLearning/Cooperative-Network-Learning}{Github Link}. Detailed descriptions of the experiments and implementation details can be found in Methods and Supplementary Note 9.

\bibliography{ref}
\section*{Acknowledgments}
This work is supported by the National Natural Science Foundation of China (Grant Nos. 61673150, 11622538, T2293771), and the New Cornerstone Science Foundation through the XPLORER PRIZE. 

\section*{Author contributions}
Q.W. and L.L. designed and coordinated the research project.
Y.H., Y.Z., Y.T., and F.Z. generated and collected the infection data.
Y.H., Y.Z., and Y.T. implemented the CNL framework and subsequent experiments.
Q.W., Y.H., and F.Z. wrote the manuscript, and L.L. reviewed the paper.
Collective contributions from all authors encompassed discussion, design, and enhancement of the CNL framework.

\section*{Competing interests}
The authors declare no competing interests.

\clearpage
\onecolumn

\begin{center}
{\Large Supplementary Information for} \\  
\vspace{0.8em}
\textbf{\Large  Cooperative Network Learning for Large-Scale and Decentralized Graphs}
\end{center}
\vspace{1em}

\setcounter{section}{0}
\setcounter{equation}{0}
\setcounter{figure}{0}
\setcounter{table}{0}
\setcounter{footnote}{0}
\makeatletter
\renewcommand{\thesection}{S\arabic{section}}
\renewcommand{\theequation}{S\arabic{equation}}
\renewcommand{\thefigure}{S\arabic{figure}}
\renewcommand{\thetable}{S\arabic{table}}

\section*{Supplementary Note 1: Three-level GNN computing process}
\addcontentsline{toc}{section}{Supplementary Note 1: Three-level GNN computing process}

\begin{figure}[!h]
\centering
\includegraphics[width=0.7\linewidth]{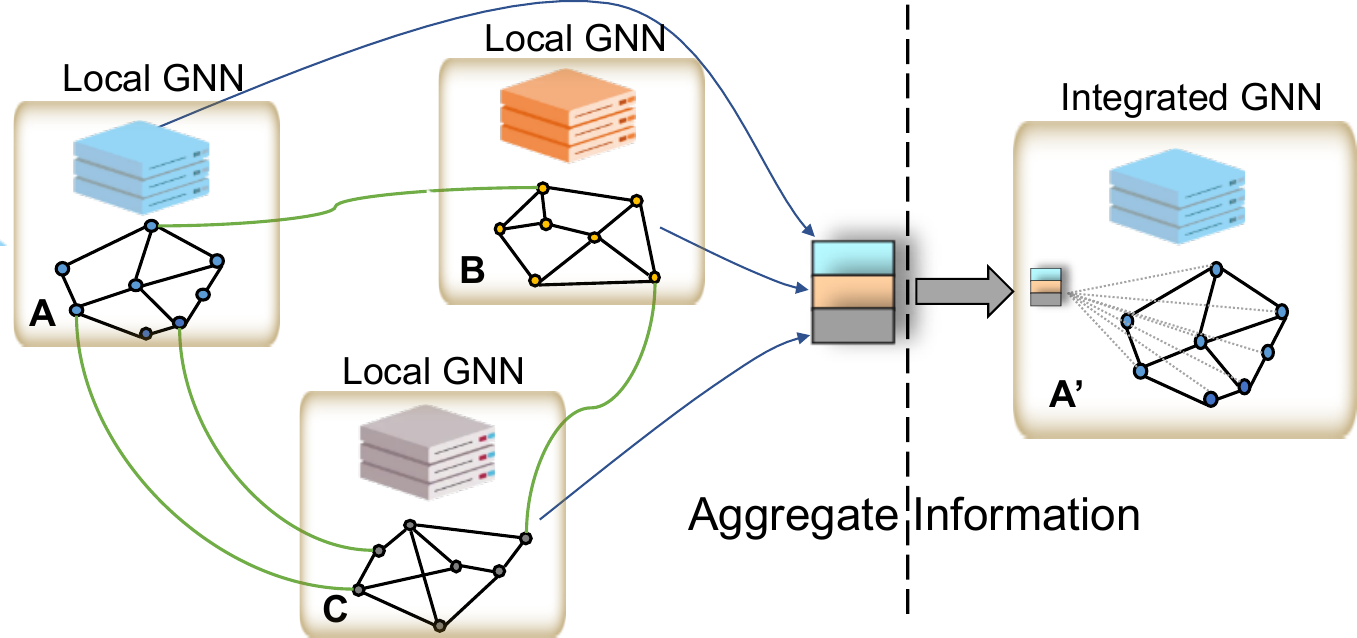}
\caption{{\bf Information aggregation under CNL framework.} Local GNN computes the embeddings of local graphs belonging to different agencies (A, B, C); Global GNN aggregates the local embeddings into global embedding for agency A; Integrated GNN sends global embedding into integrated graph A' to improve its local GNN performance.}
\label{fig:converge}
\end{figure}

The proposed CNL framework provides a reliable, fair,
privacy-preserving, and global perspective to build effective and customized models for varied tasks. In our experiments, we apply the CNL framework with GNN models (GCN \cite{GCN2017}, GAT \cite{Petar:2018gat}, and customized GNN) to two major types of tasks: contagion dynamics prediction and traditional graph tasks, i.e.,  node classification and weighted-link prediction. Contagion dynamics prediction and weighted-link prediction are continuous prediction tasks, while node classification is a discrete prediction task. We proceed to illustrate how the GNN works under the CNL framework from three perspectives, namely local, global, and integrated (as shown in Fig.~\ref{fig:converge}).

\paragraph{Local GNN}
These GNN models only have access to part of the topology and information of the network.
First, we design our local GNN model  to concatenate the node attributes, edge attributes, and task data (it varies from task to task; in contagion dynamics, it's time-series data)
to embed as $\xi_{k_i}$ for node $i$ in agency $k$. 
It is computed by $f_{local}^{emb}: \xi_{k_i} \rightarrow \mathbb{R}^{dim}$, where $dim$ is the dimension of local node embedding.
This process follows that:
\begin{equation}
\xi_{k_i} = f_{local}^{emb}(\chi_{k_i},\mathcal{b}_{k_i},\mathcal{w}_{k_i},\chi_{\mathcal{n_i}},\mathcal{B}_{n_i},\mathcal{W}_{n_i},\theta),
\end{equation}
where $\chi_{k_i}$ is node data, 
$\chi_{\mathcal{n_i}}$ is the neighbor's node data, $\mathcal{b}_{k_i}$ is node attribute, and $\mathcal{w}_{k_i}$ is edge attribute for node $i$.
The notations $\mathcal{B}_{n_i}$ and $\mathcal{W}_{n_i}$ denote the set of neighbor attributes for nodes and edges of node $i$, respectively. 
Then, we aggregate the features of the neighbors using a GNN mechanism to finally compute the outcome 
$\{\hat{\chi}_{k_i}^{t+\Delta t  },\hat{s}_{k_i}^{t+\Delta t}\}$ by $f_{local_{gnn}}$, which follows that
\begin{equation}
     \{\hat{\chi}_{k_i}^{t+\Delta t},    \hat{s}_{k_i}^{t+\Delta t}\} = f_{local_{gnn}}(\xi_{k_i}^t,\xi_{\mathcal{N}_{i}}^t,\theta),
\end{equation}
where $\hat{\chi}_{k_i}^{t+\Delta t}$ is a one-dimensional continuous variable of the continuous tasks (like contagion dynamics prediction and weighed-link prediction), and $\hat{s}_{k_i}^{t+\Delta t}$ corresponds to the probability of different discrete tasks' states such as the node classification. $\mathcal{N}_{i}$ represents the set of neighbor nodes connected to node $i$.

There are two types of loss functions for continuous and discrete tasks.
We use the mean square error (MSE) loss
function for continuous dynamics prediction:
\begin{equation}
  L_{local-MSE}(\hat{\chi}_{k_i},\chi_{k_i})=(\hat{\chi}_{k_i}-\chi_{k_i})^2,
\end{equation}
where $\hat{\chi_{k_i}}$ is the output of the continuous tasks and ${\chi_{k_i}}$ is the ground-truth.
The cross entropy (CE) loss function is applied for discrete tasks:
\begin{equation}
  L_{local-CE}(\hat{s}_{k_i},{s}_{k_i}) =-\sum_{d^l} \hat{s}_{k_i}\log({s}_{k_i}),
\end{equation}
where $\hat{s}_{k_i}$ is  the output of the discrete tasks. ${s}_{k_i}$ is the ground-truth one-hot coding and $d^l$ is the number of elements of the probability vector for the discrete state.
Hence, we can get the initial node embedding $\xi_{k_i}$ for a specific node $i$ in agency $k$.

\paragraph{Global GNN} The global graph can capture complex relationships among agencies.
Global GNN applies a GNN model on the global graph to capture the global information between agencies.
First, we establish a classic GNN model set named as GNNset, including GCN \cite{GCN2017}, GAT \cite{Petar:2018gat}, and GraphSAGE \cite{Hamilton:2017graphsage}.
Every agency has a fair right to be a computing initiator, that is, to call for other agencies to participate in global GNN computing with their own designed GNN model or a classic model selected from GNNset.
Meanwhile, each agency is connected by the data security framework of CNL, which contributes and updates its embedding according to its GNN computing graph on the global graph.
The communications of agencies are via encrypted embeddings based on CNL.
For a participated agency $k$, the initial node of global graph embedding $\Xi_k^{init} $ is aggregated from its local node embeddings as follows
\begin{equation}
    \Xi_k = \rho\left(\left\{\xi_{k_i} | \text{ if node } i \in \text{ agency } k\right\}\right).
\end{equation}
Here, $\rho$ denotes a pooling function satisfying permutation invariance, and $\xi_{k_i}$ denotes $i$-th node embedding in agency $k$. For instance, $\rho$ can be a mean function, and then the above process is simplified as $\Xi_k^{} =\sum_{i} \xi_{k_i}/I$, where $I$ denotes the number of nodes in the local graph of agency $k$.
Subsequently, $\Xi_k$ is updated by aggregating its neighbors' encrypted embeddings (information) based on CNL architecture.

Thus, $\Xi_{k}$ is encrypted as $Encryp(\Xi_{k})$, and the embeddings of $agency_k$'s neighbors are  encrypted as $Encryp(\Xi_{\mathcal{N_k}}^{})=[Encryp(\Xi_{n}), \cdots,  Encryp(\Xi_{m})]$, where $n$, $m$ are neighbors of $agency_k$.
According to Eq. (\ref{eq:global-embed}), we compute the global graph embedding  $Encryp(\Xi_{k}) \rightarrow \mathbb{R}^{dim}$ based on homomorphic encryption from CNL API ($dim$ denotes the dimension of global node embeddings) as follows
\begin{equation}
     \label{eq:global-embed}
     Encryp(\Xi_{k}) = f_{global}^{emb}(Encryp(\Xi_{k}), Encryp(\Xi_{\mathcal{N_k}}^{init}), \Theta),
\end{equation}
where $\Theta$ is the parameter.

For data privacy, we do not take into account the specific node and edge attributes of agency $k$ and its neighbors, as they have already been incorporated into the embeddings.

After that, the agency $k$ decrypts
the $Encryp(\Xi_{k})$ to obtain updated $\Xi_{k}$.
In $agency_k$, we apply a global GNN to compute the global prediction outcome  $\{\hat{X}_k^{t+\Delta t}, \hat{S}_k^{t+\Delta t}\}$ by $f_{global_{gnn}}$ as follows
\begin{equation}
     \{\hat{X}_k^{t+\Delta t}, \hat{S}_k^{t+\Delta t}\} = f_{global_{gnn}}(\Xi_k^t,\Theta).
\end{equation}
Here, $\hat{X}_k^{t+\Delta t}$ is a one-dimensional continuous variable of the continuous tasks, $\hat{S}_k^{t+\Delta t}$ corresponds to the probability of different discrete tasks' states such as the node classification.
Thus, $\Xi_k $ is further updated by the loss function from its agency. 
Also, we use the mean square error (MSE) loss
function for continuous dynamics prediction for continuous tasks:
\begin{equation}
  L_{global-MSE}(\hat{X}_k,X_k)=(\hat{X}_k-X_k)^2,
\end{equation}
where $\hat{X_k}$ is the output of the continuous tasks and ${X_k}$ is the ground-truth.
The cross entropy (CE) loss function is applied for discrete tasks for global graph:
\begin{equation}
  L_{global-CE}(\hat{S}_k,{S}_k) =-\sum_{d^g} \hat{S}_k\log({S}_k),
\end{equation}
where $\hat{S}_k$ is the output of the category of the $k$-th agency, ${S}_k$ is the ground-truth one-hot coding of $k$-th agency and $d^g$ is the number of elements of the probability vector for the categories of agencies.
%

After the loss is updated, $agency_k$ continuously updates its embedding with its neighbor nodes, and so on, for several rounds.

\paragraph{Integrated GNN} The integrated graph represents that one agency node (global view) virtually connects all nodes in its local graph. 

Integrated GNN incorporates a private model on the integrated graph to simultaneously capture global and local information.
Specifically, the private model can be either established graph learning models or customized models.
For a specific node $i$ in agency $k$, the initial node embedding is its local graph embedding $\xi_{k_i}$, and the virtual node embedding is $\Xi_k$.
Hence, the embedding of node $i$ in the integrated graph is computed as $\hat{\xi}_{k_i} \rightarrow \mathbb{R}^{dim}$, $dim$ is the dimension of integrated node features.
In mathematics, it follows that
\begin{equation}
     \hat{\xi}_{k_i} =f_{integrated}^{emb}(\xi_{k_i},\xi_\mathcal{N_{k_i}},\Xi_k,
     \hat{\theta}),
\end{equation}
where $\xi_\mathcal{N_{k_i}}$ is a set of node embedding of neighbors except the virtual node for node $i$.
According to Eq. (\ref{eq:inter-outcome}), we aggregate the features of the neighbors using an integrated GNN to finally compute the integrated prediction outcome  $\{\hat{\chi}_{k_i}^{t+\Delta t}, \hat{s}_{k_i}^{t+\Delta t}\}$ by $f_{integrated}$ as follows
\begin{equation}
     \label{eq:inter-outcome}
     \{\hat{\chi}_{k_i}^{t+\Delta t}, \hat{s}_{k_i}^{t+\Delta t}\} = f_{integrated}(\Xi_k^t,\xi_{k_i}, \xi_{\mathcal{N}_{i}}^t, \hat{\theta}).
\end{equation}
Here, $\hat{\chi}_i^{t+\Delta t}$ is a one-dimensional continuous variable of the continuous tasks, $\hat{s}_i^{t+\Delta t}$ corresponds to  the ground-truth one-hot coding and $d^i$ is the number of elements of the probability vector for the discrete tasks, and $\hat{\theta}$ is a parameter.
Note that the virtual node's final result does not need to compute.

Similarly, we use the mean square error (MSE) loss
function for continuous tasks:
\begin{equation}
  L_{integrated-MSE}(\hat{\chi}_{k_i},\chi_{k_i})=(\hat{\chi}_{k_i}-\chi_{k_i})^2,
\end{equation}
where $\hat{\chi}_{k_i}$ is the output of the continuous tasks and $\chi_{k_i}$ is the ground-truth.
The cross entropy (CE) loss function is applied for discrete tasks:
\begin{equation}
  L_{integrated-CE}(\hat{s}_{k_i},{s}_{k_i}) =-\sum_{m} \hat{s}_{k_i}\log({s}_{k_i}).
\end{equation}
Here, $\hat{s}_{k_i}$ denotes the output of the prediction results of the discrete tasks, ${s}_{k_i}$ is the ground truth one-hot coding and $d^i$ is the number of elements of the probability vector for the discrete tasks.

\section*{Supplementary Note 2: CNL framework}
\addcontentsline{toc}{section}{Supplementary Note 2: CNL framework}

The CNL framework is different from the federated learning framework in that it does not require the direct training of a target model or the explicit assignment of a model. Instead, it allows users to define the model according to their specific task. 
Basically,  the CNL framework involves each agency updating its local embeddings based on not only its own data but also data from
neighboring agencies. 
And for each task, it allows users to design integrated GNNs with customized parameters to represent the embedding of the local graph and update this local embedding based on neighbors' embeddings. 
In this way, the CNL framework generates the node embedding that integrates global information according to certain customized parameters.

In the CNL framework,  nodes are located within agencies at the local graph level, and the edges connecting them are also confined to the same agency. When an agency performs a task, it can only utilize the sub-graph and topological information that belongs to it, without access to edge information crossing into other agencies. 
From a global graph perspective, agencies themselves are represented as nodes, interconnected by edges. These edges, spanning different agencies, can be viewed as global connections between them, and are widely used in various tasks.
Through this approach, the CNL framework utilizes both global and local graph data.

\subsection*{Construction of CNL global graph}
\begin{figure}[!h]
\centering
\includegraphics[width=1\linewidth]{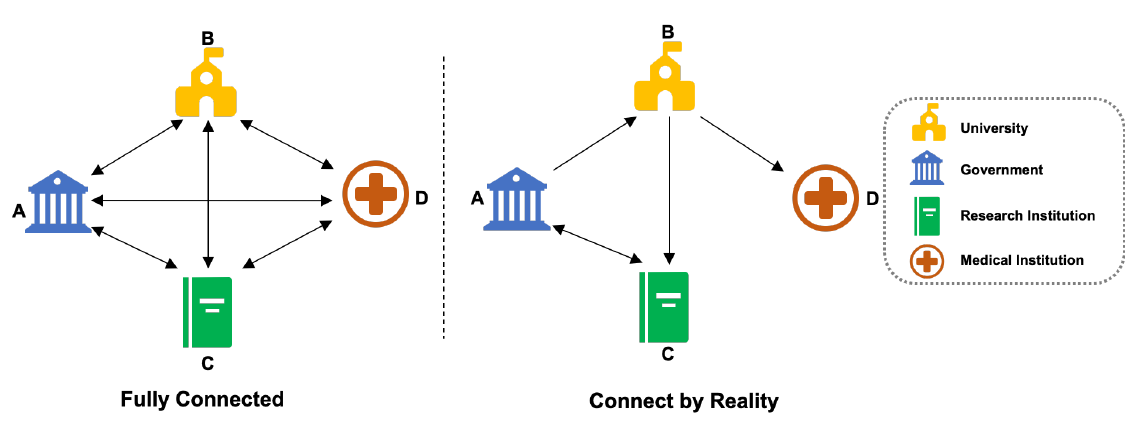}
\caption{{\bf Construction of CNL global graph.} There are two methods for connecting different agencies: 
"fully connected" means any pair of agencies are connected, which is also the default configuration of CNL, while "connected by reality" means their connections are based on actual relationships.
}
\label{fig:connectWays}
\end{figure}

In the CNL framework, there are two ways to construct the global graph: fully connected and connected by reality (see Fig. \ref{fig:connectWays} for details).

(1) "fully connected": 
Utilizing peer-to-peer networks to link all online agencies together and integrate them into the CNL framework ensures the agencies’ customized GNN model independence from practical task constraints, such as situations where agencies fail to establish a relationship with newly added agencies. 
Such a model is fit for whole downstream graph tasks, such as molecular toxicity prediction and community classification.

(2) "connected by reality": By specifying the actual relationship between a given agency and newly added online agencies, the topological structure of global nodes is determined by an agency itself, which means each agency can control which other agencies' internal activities can influence it by maintaining a neighbors list. 
Such a way is based on realistic situations: there are cross-domain edges among the agencies' internal nodes. 
This model is suitable for downstream node-level tasks, such as node classification and prediction.

\subsection*{Overall CNL framework}

\begin{figure*}[!h]
\centering
\includegraphics[width=0.8 \linewidth]{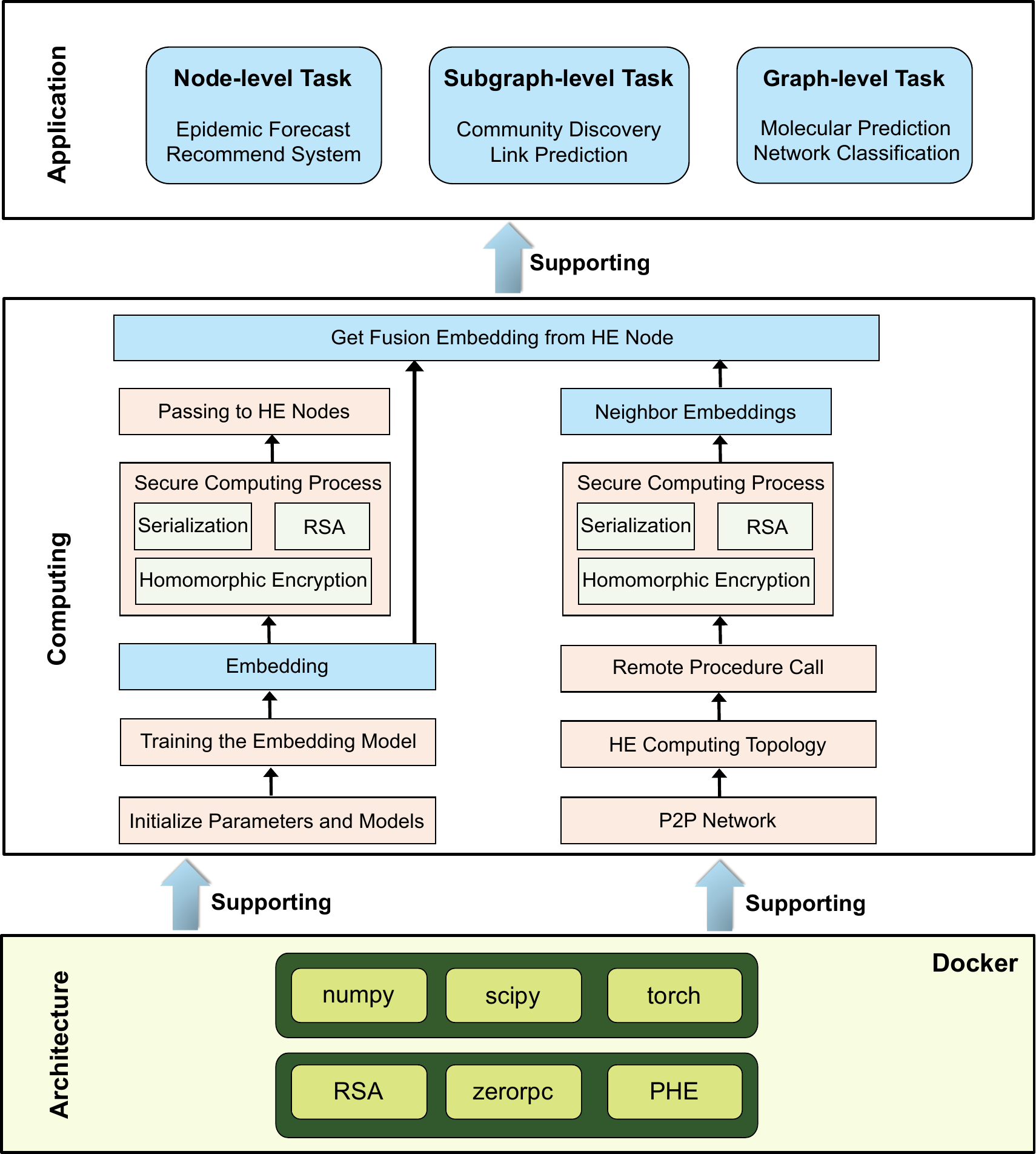}
\caption{{\bf Overview of the CNL architecture.} 
The architecture of the CNL framework can be distilled into three related modules: the architecture layer, the computing layer, and the application layer. 
The primary function of the architecture layer resides in its functional library, which adeptly orchestrates comprehensive data provisioning to underpin the entirety of the CNL ecosystem. Specifically, numpy facilitates advanced numerical computations, scipy provides a rich set of scientific algorithms for various data analysis and processing tasks, and torch empowers the system with deep learning capabilities.
At the core of the CNL service is the computational layer, which allows the integration of information from neighboring agencies and the acquisition of fusion embeddings, while maintaining strict privacy safeguards. The resulting fusion embeddings are subsequently channeled into various downstream applications, thereby underscoring CNL's adaptive potential. The computational layer encompasses tasks such as training embedding models, generating embeddings, and ensuring data security through Serialization, RSA, and Homomorphic Encryption. It relies on communication with neighboring nodes via a P2P network to obtain the Fusion Embedding.
The versatility of CNL is manifested in its ability to accommodate node-level, subgraph-level, and graph-level tasks, and ensure customization of downstream models.
}
\label{fig:Overall_framework}
\end{figure*}

The comprehensive processing architecture of the CNL model can be distilled into three related modules: the architecture layer, the computing layer, and the application layer, as illustrated in Fig. \ref{fig:Overall_framework}. 
Specifically, the functional library within the architecture layer plays a role in orchestrating the holistic data support for the entire CNL system. 
At the core of the CNL service, the computational layer facilitates the integration of information from neighboring agencies and generates fusion embeddings while ensuring data privacy.
These resulting fusion embeddings can then be channeled into various downstream applications. 
The versatility of CNL is manifested in its ability to accommodate node-level, subgraph-level, and graph-level tasks, and ensure customization of downstream models.

As shown in the middle panel of Fig. \ref{fig:Overall_framework}, Within the computing layer, the process unfolds as follows: When an agency requires CNL service, it begins by initializing its local model and parameters. Then it trains the embedding model to obtain the local embedding.
Throughout this process, the CNL framework encourages the sharing of graph embedding among agencies; 
this means that each agency can obtain embeddings from its neighboring agencies, enabling the construction of an integrated model that surpasses the capabilities of its local model. 
Concurrently, each agency is supposed to share its local embedding with the homomorphic encryption computing node for the requisition from other agencies. 

Specifically, as depicted in the right panel of the computing layer in Fig. \ref{fig:Overall_framework}, the CNL is based on the P2P (Peer-to-Peer) technological concept, ensuring voluntary and equitable participation of data owners in the CNL computational graph. In the left panel, it encompasses the initialization and training process of embedding models, with data securely transmitted to HE (Homomorphic Encryption) nodes within the same security module.
The communication within CNL is facilitated by Remote Procedure Call based on the TCP/IP protocol. 
Subsequently, CNL leverages multifaceted technologies, namely serialization, RSA, and Homomorphic encryption, to ensure a secure computing process. 
Ultimately, the agency that requires CNL service can acquire the neighbor embeddings, which are then incorporated with its local embedding to engender the fusion embedding.


\subsection*{Model training and acquisition}
The information transmitted among agencies is the embedding of the agency, which comes in various types, such as GCN \cite{GCN2017}, GAT \cite{Petar:2018gat}, and GraphSAGE \cite{Hamilton:2017graphsage}. 
Since the performance of different embeddings can vary significantly, it is imperative for the CNL framework to train a model tailored to the specific task at hand, and share the obtained embeddings with other agencies based on the secure computing process. 

When an agency has tasks, it first assigns task datasets, embedding models, and other hyper-parameters according to the task parameters list.  Then it transmits the data to its neighboring agencies. Finally, the neighboring agency chooses and executes the embedding of the model and calculates the embedding of the agency itself according to the parameters list.

The parameters list includes arguments defined using the args class, 
a standard module in Python for command-line parsing. Within this list, argsparse specifies the required dataset for the task, which is represented as an input with an indefinite argument list and default values. Additionally, it allows for the specification of the embedding model, its structure, and hyper-parameters related to training the embedding model.

\section*{Supplementary Note 3: Data security framework}
\addcontentsline{toc}{section}{Supplementary Note 3: Data security framework}

\begin{figure}[!ht]
\centering
\includegraphics[width=0.65\linewidth]{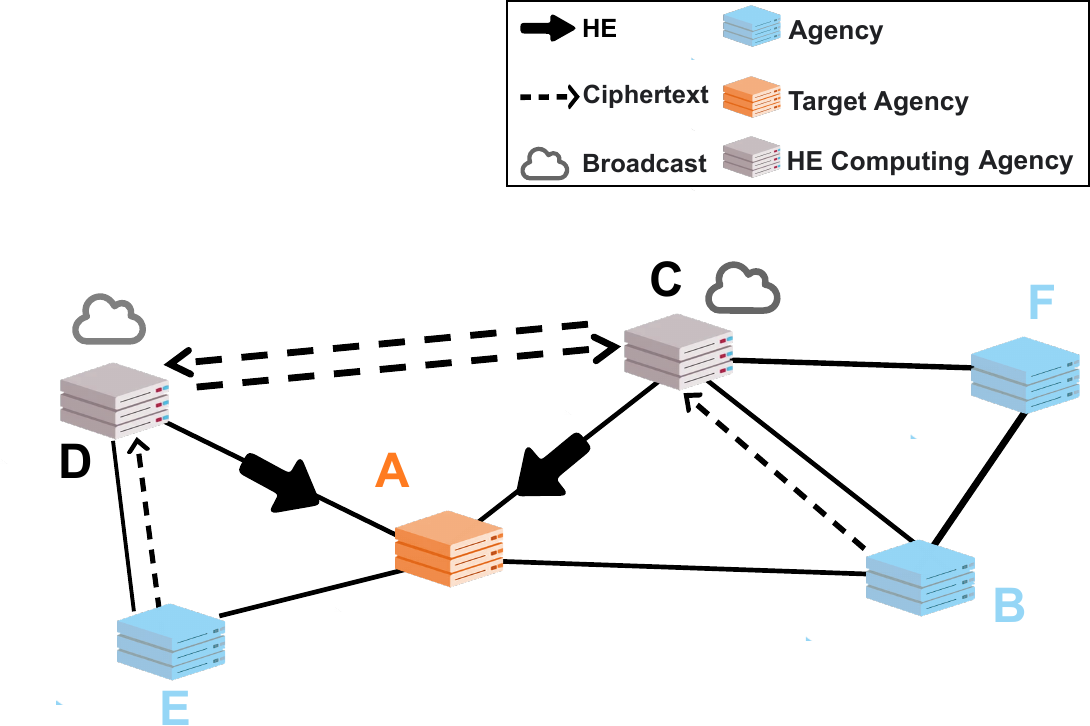}
\caption{{\bf Computing security of CNL framework.} HE computing agencies $C$ and $D$ generate ciphertext based on target agency $A$'s public key and transfer the summed homomorphic ciphertext to $A$. Here, HE computing agencies are selected randomly and the granularity of the encryption can be adjusted by setting different numbers of computing agencies. Each agency can send its own encrypted embedding to any number of HE computing agencies.}
\label{fig:HE}
\end{figure}

At the initialized stage of a given task, each agency generates a group of homomorphically encrypted keys (public keys) while sharing the same task configuration. Then, each agency passes its public key to all its neighbors. 
In Fig. \ref{fig:HE}, agency $A$ (the Target Agency) wants to update its nodes' state. It will first choose $n$ neighbor nodes randomly as computing centers for the homomorphic encryption algorithm. Here, agencies $C$ and $D$ are chosen as the HE computing agencies (Note that 
the number of computing agencies can be adjusted to control the encryption granularity).
Agency $A$ then notifies all its neighbors of the selected HE compute agencies, and these neighbors will choose to send their homomorphically encrypted embedding to some or all of the chosen HE compute agencies (0, 1, or more),  where the public key is provided by $A$.
Each selected HE computing agency collects and sums up the received embedding, and then broadcasts the summed homomorphic ciphertext file to all its neighboring agencies (including agency $A$ ).

To avoid busy waiting and improve efficiency, we design the CNL framework as follows: after an agency sends an embedding task command and configuration to its neighboring agencies, these agencies will choose a dataset and decide the hyper-parameters of the model according to the args parameters list. 
Then, they start an independent calculating thread to train the embedding model and obtain the embedding. 
After the computation ends, the computing node passes the result to those selected HE computing agencies.
Finally, agency $A$ checks with and queries the HE computing agencies to confirm whether the homomorphic encryption operation is finished.

Agency A, which has downstream task requirements, utilizes the neighbor embedding request and awaits the completion of the neighbor embedding computation. When the computation is completed, the HE computing agencies transfer the homomorphically encrypted ciphertext to $A$ and destroy its own message. The reliability of the passing process is guaranteed by remote procedure call (RPC) based on Transmission Control Protocol (TCP).

\section*{Supplementary Note 4: Cooperative network node service}
\addcontentsline{toc}{section}{Supplementary Note 4: Cooperative network node service}

\begin{figure*}[!h]
\centering
\includegraphics[width=0.8 \linewidth]{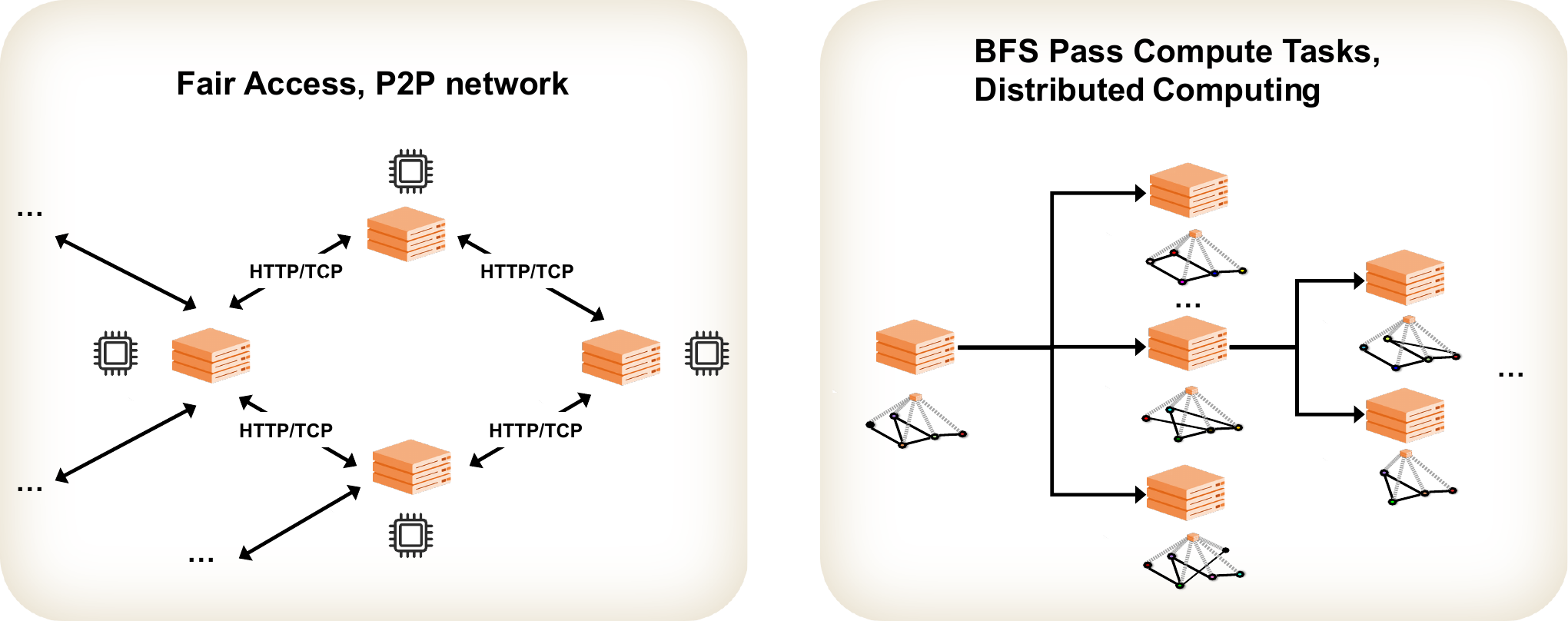}
\caption{{\bf Technical implementation architecture of CNL.} 
The CNL is based on the P2P (Peer-to-Peer) technological concept, ensuring voluntary and equitable participation of data owners in the CNL computational graph. The communication within CNL is facilitated by RPC (Remote Procedure Call) based on the TCP/IP protocol. Additionally, when a task needs to be executed, instructions are propagated throughout the entire computational graph in a breadth-first traversal manner, starting from the source node. Computation and propagation of embedded calculations are carried out on different devices within the network.
}
\label{fig:network}
\end{figure*}

Cooperative network node service (CNNS) will automatically read the configuration file, perform necessary initialization, and open the corresponding port for the call. As illustrated in Fig.~\ref{fig:network}, GNNs are connected to the CNL network through a P2P access method, and service instructions propagate within the CNL network using a BFS approach \cite{bundy1984breadth}. CNNS relies on Python 3.8 and includes third-party libraries such as Torch and Torch-Geometric.
When services are enabled on a global node, the node is regarded as "online", and the online nodes form a topology from a global perspective. 
The CNNS node receives the task, propagates the same command to its neighbors, and synchronizes tasks between global nodes by setting the \textit{task\_iter} round.

In node-level tasks like pandemic prediction, the pandemic data in different regions are linked due to cross-domain interactions, leading to correlated data across regions. The spread in both cross-domain nodes and different regions is influenced by their neighboring areas.

In link-level tasks, the CNL processing workflow is similar to that of node-level tasks. Therefore, the CNL framework can improve the performance of downstream tasks by integrating the topological structure between regions via the neighbors' embedding.

\textbf{The Cooperative Network Learning Process: }

In this section, we outline the operation and utilization process of CNL:

(1) Installation of CNNS service:
Install the CNNS service by utilizing a Docker image or directly downloading the code repository files.

(2) Configuration of CNNS service:
Configure the CNNS service by specifying the data file directory, communication ports, neighbor addresses, and other relevant information.

(3) Launching the CNNS service:
Once the CNNS service is properly configured, launch it. The corresponding ports will be opened upon successful startup, allowing external access to the CNNS service.

(4) Invoking the CNNS service for downstream tasks:
To utilize the CNNS service for downstream tasks, invoke it as needed. Depending on the requirements of the specific task, the CNNS service can interact with neighboring CNNS instances, forming a global graph structure that incorporates information from the network's participants.

\section*{Supplementary Note 5: Preliminary for prediction tasks}
\addcontentsline{toc}{section}{Supplementary Note 5: Preliminary for prediction tasks}

\subsection*{Contagion dynamics prediction}

We formulate the contagion dynamics prediction problem as a graph-based propagation model.  We have a total of $N$ regions (e.g., cities or states). We denote the historical cases data $\left[x_1,\cdots,x_t\right]$ as training data, where $x_\tau\in \mathbb{R}^N$ represents the observed case values of $N$ regions at time $\tau$. 

Our goal is to predict the future case value, i.e. $x_{t+h}$, where $h$ is a fixed horizon with respect to different tasks (e.g., short- or long-term prediction). For every task, we use $\left[x_{t-T+1},\cdots,x_t\right] \in \mathbb{R}^{N\times T}$ for a specific look-back window $T$ to predict $x_{t+h}$.

We compare our customized contagion dynamics prediction model to several state-of-the-art (SOTA) methods: Autoregressive \cite{AR2015dynamic}, LSTnet \cite{LSTnet}, ST-GCN \cite{ST-GCN}, EpiGNN \cite{EpiGNN}, ColaGNN \cite{ColaGNN} and CNNRNN-Res \cite{CNNRNN-Res}. The comprehensive introduction to these models is deferred to Supplementary Note 7.

\subsection*{Node classification}
Node classification is a fundamental task in network analysis and graph representation learning, aiming to predict the labels or categories of nodes in a network based on their attributes and network structure. This task has significant applications in various domains, including social network analysis, biological network analysis, and recommendation systems.
Notably, various GNN models have achieved outstanding performance in node classification tasks, such as GCN \cite{GCN2017}, GAT \cite{Petar:2018gat}, and GraphSAGE \cite{Hamilton:2017graphsage}.

\subsection*{Link prediction}
In this paper, link prediction specifically refers to link (edge) weight prediction, entailing the estimation or prediction of numerical values associated with edges within a graph. 
In weighted graphs, edges are assigned values that indicate the strength, proximity, similarity, or other quantitative metrics between the connected nodes. The goal of edge weight prediction is to infer or predict these values based on various graph characteristics, node attributes, or other available information.

Edge weight prediction has numerous applications in various domains. For instance, social networks can be used to estimate the strength of relationships between individuals, helping in tasks like friend recommendations or influential connection identification\cite{ISMnet}. In transportation networks, it can assist in predicting travel time or distance between locations. In biological networks, it can aid in understanding protein-protein interactions or gene regulatory networks.

\clearpage
\section*{Supplementary Note 6: Dataset description}
\addcontentsline{toc}{section}{Supplementary Note 6: Dataset description}
\label{SI:data}

We aggregate the public datasets and our self-generated datasets to this link: \href{https://drive.google.com/drive/folders/1wTwpj3XzqzySTIDDew9LVkb7_tgiwypU?usp=sharing}{Google Drive Link}

\subsection*{Synthetic contagion dynamics}

We utilize two generative models, namely the Erd{\H{o}}s–R{\'e}nyi (ER) \cite{ER1960} network and Barab{\'a}si–Albert (BA) \cite{BA1999} network, to construct the synthetic network.
Each of these datasets consists of 10,000 nodes and 50,000 edges.
Additionally, to simulate the spreading process on these synthetic graphs, we utilize two classical spreading models: Susceptible-Infected-Susceptible and Susceptible-Infected-Recovered. 
Following are the specific details of these two spreading models.

\begin{itemize}
\item \textbf{Susceptible-Infected-Susceptible model  (SIS).}
The SIS \cite{barabasi2013network} spreading model assumes that individuals have two states: susceptible and infected.  For a given network $G$, a susceptible individual $i$ can be infected by its neighbor with an infection rate of $\beta$, and the infected individual can turn to the susceptible state with a probability of $\mu$.  The spreading process will stop when it reaches a steady state.

\item \textbf{Susceptible-Infected-Recovered model  (SIR).}
The SIR \cite{barabasi2013network, morone2015influence} spreading model assumes that individuals have three states: susceptible, infected, and recovered. For a given network $G$, a susceptible individual $i$ can be infected by its neighbor with an infection rate of $\beta$, the infected individual can recover with a probability of $\mu$, and those individuals in the recovered state will not be infected again. The spreading process will stop when no individuals are in the infected state.
\end{itemize}

All datasets are split into train/validation/test sets with a ratio of 50\%/20\%/30\% in chronological order. The validation data is used to determine the number of epochs needed to prevent overfitting.

\subsection*{Empirical contagion dynamics}

Empirical experiments are carried out on four epidemic-related datasets, and their data statistics are summarized in Table \ref{tab:contagion_data_statistics}, where SD denotes standard deviation. The graph adjacency matrices are constructed based on geographical adjacency information. Furthermore, all datasets are partitioned into train, validation, and test sets in chronological order, with a ratio of 50\%, 20\%, and 30\%, respectively.  The validation data is used to determine the number of epochs that should be run to avoid overfitting.

\begin{itemize}
    \item \textbf{US-States}. This dataset collects data about influenza from the Centers for Disease Control Prevention (CDC) \cite{data:CDC}, which includes the number of patient visits for ILI (positive cases) per week and per state in the US from 2010 to 2017. We retain 49 states in this dataset after deleting Florida due to missing data. 
    \item \textbf{US-Regions}. This dataset is the ILINet portion of the US-HHS (Department of Health and Human Services) dataset \cite{data:CDC}, and it contains weekly influenza activity levels for ten HHS regions on the US mainland from 2002 to 2017.
    \item \textbf{Spain}. This dataset is the daily COVID-19 cases from 20 February 2020 to 20 June 2020 for the 35 administrative NUTS3 areas in Spain \cite{dataset_spain}.
    \item \textbf{US COVID-related retweets.} This dataset is obtained from the paper~\cite{banda_juan_m_2020_3757272}, which includes a large-scale list of COVID-related tweets~ [\url{https://github.com/thepanacealab/covid19_twitter}] since the outbreak of COVID-19. We extract US user tweets from March 23, 2020, to April 24, 2020, which includes 48 states and 210 edges. Specifically, we (\romannumeral1) extract the tweets that are joined by US users; (\romannumeral2) categorize these users according to their location.  (\romannumeral3) count state-specific retweets every $6$ hours; (\romannumeral4) create state-specific tweet count time series data.
\end{itemize}

\begin{table}[!ht]
\centering
\caption{Statistics of contagion dynamics datasets.}
\begin{threeparttable}
\begin{tabular}{lccccccc}
\toprule
Datasets   & Field   & Regions & Length & Min & Max   & Mean & SD   
\\ \midrule
US-States  & Influenza & 49      & 360    & 0   & 9716  & 223  & 428  \\
US-Regions & Influenza & 10      & 785    & 0   & 16526 & 1009 & 1351 \\
Spain      & COVID-19  & 35      & 122    & 0   & 4623  & 38   & 269  \\ 
Twitter      & Social platform  & 48      & 132  & 1   &  47930 &  1789  & 4693  \\ \bottomrule
\end{tabular}
\end{threeparttable}
\label{tab:contagion_data_statistics}
\end{table}

\subsection*{Node classification}
We perform the node classification experiments on two homogeneous graphs (i.e., Cora \cite{data:Cora} and PubMed \cite{data:PubMed}) and two heterogeneous graphs (i.e., Texas and Wisconsin  \cite{data:Pei2020Geom-GCN}).

The Cora and PubMed datasets are two citation networks where nodes represent scientific papers, and edges represent citations between these papers. Each paper is associated with a word vector indicating the presence of specific words from a dictionary. The task is to classify the documents into predefined categories.
Both Texas and Wisconsin are part of the WebKB dataset, which contains web pages from computer science departments of various universities. The nodes in these graphs represent web pages, and the edges represent hyperlinks between them. The task is to classify each webpage into one of several predefined categories, such as faculty, student, project, etc.

Adjacent nodes in a homogeneous graph tend to share the same label, while the opposite holds in heterogeneous graphs \cite{homo2020}. 

The basic statistics of these datasets are reported in Table \ref{tab:node_statistics}.
The homophily level \cite{homo2020} of network $G$ is measured by $Homophily(G)=\frac{\left|\left\{(u, v):(u, v) \in E \wedge y_v=y_u\right\}\right|}{m}$, where $y_v$ is the label of nodes $v$ and $m = |E|$ denotes the number of edges.

\begin{table}[htbp]
\centering
\caption{Statistics of node classification datasets.}
\begin{tabular}{cccccccc}
\toprule
Network   & $\#$Classes & $\#$Features  & $\#$Nodes  & $\#$Edges    &$\left \langle k \right \rangle$ & Homophily\\
\midrule
Cora      & 7       & 1433  & 2708  & 5278  &3.90   & 0.8099\\
PubMed    & 5       & 500   & 19717 & 44324 &4.50   & 0.8023\\
Texas     & 5       & 1703  & 183   & 279   &3.05   & 0.0871\\
Wisconsin & 5       & 1703  & 251   & 450   &3.56   & 0.1921\\
\bottomrule
\end{tabular}
\label{tab:node_statistics}
\end{table}

\subsection*{Link prediction}
We conduct link prediction experiments on the Ciao dataset \cite{tang2012mtrust}, a heterogeneous graph dataset consisting of 12,375 user nodes and 106,797 product nodes, with a total of 237,350 edges. The products are categorized into 28 distinct categories. The clustering coefficient of the graph is calculated to be 0.1969.

\section*{Supplementary Note 7: Introduction to comparison methods}
\addcontentsline{toc}{section}{Supplementary Note 7: Introduction to comparison methods}
\label{SI:methods}

\subsection*{ Methods for contagion dynamics prediction}
We compare our customized contagion dynamics prediction model to several state-of-the-art (SOTA) methods and their variations, as listed below.
\begin{itemize}
\item \textbf{Autoregressive (AR) \cite{AR2015dynamic}.} Autoregressive models are widely utilized in time series forecasting, where the future outcomes are modeled as a linear combination of past data points. We train an individual autoregressive model for each location (node), with no data or parameters being shared between nodes.

\item \textbf{LSTnet \cite{LSTnet}.} This model employs CNN and RNN to identify short-term local dependencies among variables and uncover long-term trends in time series data.

\item \textbf{ST-GCN \cite{ST-GCN}.} A deep learning framework, initially developed for traffic prediction, integrates graph convolution and gated temporal convolution through spatio-temporal convolutional blocks.

\item \textbf{EpiGNN \cite{EpiGNN}.} This model introduces a transmission risk encoding module to capture local and global spatial effects for each region. It also considers transmission risk, geographical dependencies, and temporal information to better explore spatial-temporal dependencies.

\item \textbf{ColaGNN \cite{ColaGNN}.} This model introduces a new dynamic adjacency matrix, leveraging cross-location attention scores to identify directed spatial effects. Additionally, a multi-scale dilated convolution layer is adopted on time series to capture both short and long-term patterns.

\item \textbf{CNNRNN-Res \cite{CNNRNN-Res}.} A deep learning framework for epidemiological prediction problems that combines CNN, RNN, and residual links. It fuses information from different locations using CNN.

\end{itemize}

\subsection*{Methods for traditional graph tasks }

Graph neural networks (GNNs) have emerged as a powerful framework for analyzing and learning from graph-structured data. Unlike traditional neural networks designed for grid-like data, GNNs are specifically tailored to capture and exploit the underlying structural information present in graphs\cite{ISMnet}. This makes them particularly well-suited for tasks involving network analysis, such as node classification, link prediction, and contagion dynamics prediction.

Basically, GNNs operate by iteratively updating node representations based on the information from their neighboring nodes, allowing for information propagation and aggregation across the graph. This process enables GNNs to effectively capture complex dependencies and relationships between nodes, leveraging both the local neighborhood and the global graph structure.
The versatility of GNNs lies in their ability to incorporate both node features (attributes) and graph structure during the learning process. This allows GNNs to effectively model and capture the intricate interplay between node characteristics and their connections within the graph\cite{HiGCN}.

Various GNN models can be incorporated into our CNL framework. Here are some traditional GNN models employed in this study.

\begin{itemize}
\item \textbf{GCN \cite{GCN2017}.} The GCN model introduces a localized graph convolution operation that aggregates information from a node's immediate neighborhood. By propagating and updating node features iteratively, GCN captures both local and global structural information in the graph. 

\item \textbf{GraphSAGE \cite{Hamilton:2017graphsage}.} GraphSAGE model is a graph representation learning algorithm that aims to generate embeddings for nodes in a graph. 
Specifically, it learns node embeddings by sampling and aggregating information from the neighborhood nodes, using a neural network architecture that can generalize to unseen nodes. By considering the collective information from the graph, GraphSAGE can produce powerful node representations that can be used for various downstream tasks, such as node classification and link prediction.

\item \textbf{GAT \cite{Petar:2018gat}.} 
The GAT model introduces an attention mechanism to capture the importance of different nodes in the neighborhood aggregation process. By assigning learnable attention weights to neighbors, GAT allows for adaptive and context-dependent aggregation of information. This model has demonstrated superior performance in various tasks.

\end{itemize}

For these baseline implementations, we employ the source codes released online and adopt the best parameter settings for each method.

\clearpage
\section*{Supplementary Note 8: Supplementary experimental results}
\addcontentsline{toc}{section}{Supplementary Note 8: Supplementary experimental results}

\subsection*{ Results of contagion dynamics on synthetic data}

To assess the efficacy of the CNL framework, we conduct contagion dynamics prediction on Erd{\H{o}}s–R{\'e}nyi (ER) \cite{barabasi2013network} and Barab{\'a}si–Albert (BA) \cite{BA1999} graphs with two spreading models, namely Susceptible-Infected-Recovered (SIR) and Susceptible-Infected-Susceptible (SIS) model \cite{barabasi2013network}. 
The results are reported in Tables \ref{tab: empirical_generated_SIR} and \ref{tab: empirical_generated_SIS}, respectively.

\begin{table*}[!h]
\centering
\caption{Empirical prediction performance on contagion dynamics of local, integrated and centralized GNNs in datasets generated by SIR simulation.}
\begin{tabular}{clcccccc}
\toprule
\multirow{2}{*}{Performance} & \multicolumn{1}{c}{\multirow{2}{*}{Metric}} & \multicolumn{3}{c}{ER network}     & \multicolumn{3}{c}{BA network}     \\ \cmidrule(r){3-5} \cmidrule(r){6-8}
& \multicolumn{1}{c}{}                        & Local   & Integrated  & Centralized & Local   & Integrated  & Centralized \\ \midrule
\multirow{2}{*}{Agency-A}      
& RMSE & 26.2458     & 3.4419    & 20.5112       & 7.5427     & 2.7740         &27.4972   \\
& PCC  & 0.1097     & 0.2306    & 0.0541       & 0.1011      &0.6870       &0.3093 \\
\multirow{2}{*}{Agency-B}      
& RMSE & 9.0174     & 5.3602     & 20.2711    & 28.1504    &9.0049        &30.8724 \\
& PCC  & 0.0130      & 0.0756     & 0.2401     & 0.4424   & 0.6900       &0.3686\\
\multirow{2}{*}{Agency-C}      
& RMSE & 14.3164     & 4.6624     & 21.9201    & 1.8303    & 9.0049        &27.8608 \\
& PCC  & 0.3097     & 0.3957     & 0.0418     & 0.0149   & 0.0690       &0.0327  \\
\multirow{2}{*}{Agency-D}      
& RMSE & 3.7507     & 3.7506     & 21.6187    & 1.6851    & 2.1252        &26.4755 \\
& PCC  & 0.6440     & 0.6440     & 0.2698    & 0.5209   &0.5043       &0.0260  \\
\multirow{2}{*}{Agency-E}      
& RMSE & 20.8500     & 1.5295    & 21.0993    & 13.2831     &9.5826     &28.4860    \\
& PCC  & 0.0116      & 0.4924    & 0.0606     & 0.0635     &0.2910    &0.0810  \\ \midrule
\multirow{2}{*}{Multi-agency} 
& RMSE & 56.1018     & 3.8522    & 9.2095    & 13.4707     &6.8751         &28.1036    \\
& PCC  & 0.0352      & 0.1467    & 0.2401     & 0.0567   &0.1765       &0.1257  \\ \bottomrule
\end{tabular}
\label{tab: empirical_generated_SIR}
\end{table*}

\begin{table*}[!h]
\centering
\caption{Empirical prediction performance on contagion dynamics of local, integrated and centralized GNNs in datasets generated by SIS simulation.}
\label{tab: empirical_generated_SIS}
\begin{tabular}{clcccccc}
\toprule
\multirow{2}{*}{Performance} & \multicolumn{1}{c}{\multirow{2}{*}{Metric}} & \multicolumn{3}{c}{ER network}     & \multicolumn{3}{c}{BA network}     \\ \cmidrule(r){3-5} \cmidrule(r){6-8}
& \multicolumn{1}{c}{}                        & Local   & Integrated  & Centralized & Local   & Integrated  & Centralized \\ \midrule
\multirow{2}{*}{Agency-A}      
& RMSE & 5.8340     & 5.4556    & 5.5843       & 6.1281     & 6.1225         &6.4672   \\
& PCC  & 0.9826     & 0.9836    & 0.9825       & 0.9817     & 0.9827       &0.9822 \\
\multirow{2}{*}{Agency-B}      
& RMSE & 6.0279     & 5.5896     & 5.2428     & 9.6197      & 8.9832        &8.8166 \\
& PCC  & 0.9754     & 0.9762     & 0.9752     & 0.9458      & 0.9515       &0.9551\\
\multirow{2}{*}{Agency-C}      
& RMSE & 6.7438     & 6.5141     & 6.9949     & 6.8315    & 7.3805       &6.1807 \\
& PCC  & 0.9003     & 0.9045     & 0.8980     & 0.9064   & 0.8771       &0.9057  \\
\multirow{2}{*}{Agency-D}      
& RMSE & 5.0544     & 4.8616     & 4.8014    & 5.6527    & 5.2984        &5.1322 \\
& PCC  & 0.9735     & 0.9732     & 0.9735    & 0.9729    & 0.9736       &0.9732  \\
\multirow{2}{*}{Agency-E}      
& RMSE & 5.9905     & 5.7909    & 5.6337     & 5.8057     &5.8794     &6.0772    \\
& PCC  & 0.9890     & 0.9890    & 0.9889     & 0.9868     &0.9863    &0.9821  \\ \midrule
\multirow{2}{*}{Multi-agency} 
& RMSE & 6.1501     & 5.7912    & 5.6036     & 6.7913     &6.6771         &5.8032    \\
& PCC  & 0.9720     & 0.9807    & 0.9821     & 0.9746     &0.9762       &0.9868  \\ \bottomrule
\end{tabular}
\end{table*}

Generally, in addition to utilizing local information, the integrated models are provided with supplementary embeddings (local and global embeddings), leading to improved performance compared to local models.

\subsection*{ Results of contagion dynamics on empirical data}

In our customized model, diverse GNN models can be integrated to examine the dynamic spread of contagion across various regions, concurrently integrating embeddings provided by other agencies.
We show the results obtained using GAT \cite{Petar:2018gat} and GraphSAGE \cite{Hamilton:2017graphsage} as spatial information processing modules in Tables \ref{tab:empirical_GAT} and \ref{tab:empirical_SAGE}, respectively.

Tables \ref{tab:empirical_GAT} and \ref{tab:empirical_SAGE} summarize the single-agency and multi-agency collective performance of various models (i.e., Local, Integrated, and Centralized models) in terms of RMSE and PCC. 
The results suggest that integrated models generally outperform local models.

\begin{table*}[!ht]
\centering
\caption{Empirical prediction performance on contagion dynamics of local, integrated, and centralized GNNs in real-world datasets incorporating CNL framework and GAT layer. The integrated models with improved performance compared to local models are highlighted in bold.}
\label{tab:empirical_GAT}
\begin{tabular}{cccccccccc}
\toprule
\multirow{2}{*}{Dataset}   
& \multirow{2}{*}{Model} 
& \multicolumn{2}{c}{Agency-A}                       
& \multicolumn{2}{c}{Agency-B}                       
& \multicolumn{2}{c}{Agency-C}                       
& \multicolumn{2}{c}{Multi-agency}                         \\
\cmidrule(r){3-4} \cmidrule(r){5-6} \cmidrule(r){7-8} \cmidrule(r){9-10}
                            &                       & \multicolumn{1}{l}{RMSE} & \multicolumn{1}{l}{PCC} & \multicolumn{1}{l}{RMSE} & \multicolumn{1}{l}{PCC} & \multicolumn{1}{l}{RMSE} & \multicolumn{1}{l}{PCC} & \multicolumn{1}{l}{RMSE} & \multicolumn{1}{l}{PCC} \\ 
\midrule
\multirow{3}{*}{US-States} & Local                 & 113                      & 0.889                   & 381                      & 0.781                   & 143                      & 0.591                   & 266                      & 0.805                   \\
                           & Integrated            & \bf{107}                      & \bf{0.897}                   & \bf{342}                      & \bf{0.834}                   & 143                      & 0.591                   & \bf{242}                      & \bf{0.849}                   \\
                           & Centralized           & 110                      & 0.891                   & 315                      & 0.852                   & 126                      & 0.724                   & 224                      & 0.865                   \\ \midrule
\multirow{3}{*}{US-region} & Local                 & 898                      & 0.817                   & 1095                     & 0.751                   & 511                      & 0.799                   & 915                      & 0.768                   \\
                           & Integrated            & 898                      & 0.817                   & 1095                     & 0.751                   & \bf{382}                      & \bf{0.909}                   & \bf{896}                      & \bf{0.779}                   \\
                           & Centralized           & 795                      & 0.854                   & 1092                     & 0.748                   & 379                      & 0.860                   & 875                      & 0.789                   \\ \midrule
\multirow{3}{*}{Spain}     & Local                 & 248                      & 0.236                   & 19                       & 0.618                   & 7                        & 0.053                   & 192                      & 0.236                   \\
                           & Integrated            & \bf{243}                      & \bf{0.277}                   & 19                       & 0.618                   & \bf{6}                        & \bf{0.002}                   & \bf{188}                      & \bf{0.275}                   \\
                           & Centralized           & 220                      & 0.394                   & 18                       & 0.546                   & 62                       & -0.058                  & 172                      & 0.386                   \\ \midrule
\multirow{3}{*}{Twitter}   & Local                 & 2070                     & 0.962                   & 1453                     & 0.957                   & 687                      & 0.870                   & 1455                     & 0.865                   \\
                           & Integrated            & \bf{1959}                     & \bf{0.964}                   & \bf{1430}                     & \bf{0.960}                   & 687                      & 0.870                   & \bf{1398}                     & \bf{0.876}                   \\
                           & Centralized           & 1872                     & 0.962                   & 1381                     & 0.963                   & 668                      & 0.876                   & 1341                     & 0.959                   \\ 
\bottomrule
\end{tabular}
\end{table*}

\begin{table}[!ht]
\centering
\caption{Empirical prediction performance on contagion dynamics of local, integrated, and centralized GNNs in real-world datasets incorporating CNL framework and GraphSAGE layer.  The integrated models with improved performance compared to local models are highlighted in bold.}
\label{tab:empirical_SAGE}
\begin{tabular}{cccccccccc}
\toprule
\multirow{2}{*}{Dataset}    
& \multirow{2}{*}{Model} 
& \multicolumn{2}{c}{Agency-A}                       
& \multicolumn{2}{c}{Agency-B}                       
& \multicolumn{2}{c}{Agency-C}                      
& \multicolumn{2}{c}{Multi-agency} \\ 
\cmidrule(r){3-4} \cmidrule(r){5-6} \cmidrule(r){7-8} \cmidrule(r){9-10}
                            &                       & \multicolumn{1}{l}{RMSE} & \multicolumn{1}{l}{PCC} & \multicolumn{1}{l}{RMSE} & \multicolumn{1}{l}{PCC} & \multicolumn{1}{l}{RMSE} & \multicolumn{1}{l}{PCC} & \multicolumn{1}{l}{RMSE} & \multicolumn{1}{l}{PCC} \\ \midrule
\multirow{3}{*}{US-States}  & Local                 & 110                      & 0.893                   & 332                      & 0.835                   & 130                      & 0.674                   & 235                      & 0.851                   \\
                            & Integrated            & \bf{105}                      & \bf{0.903}                   & 332                      & 0.835                   & 130                      & 0.674                   & \bf{234}                      & \bf{0.852}                   \\
                            & Centralized           & 104                      & 0.903                   & 306                      & 0.860                   & 127                      & 0.703                   & 218                      & 0.872                   \\ \midrule
\multirow{3}{*}{US-Regions} & Local                 & 983                      & 0.698                   & 1193                     & 0.687                   & 458                      & 0.801                   & 983                      & 0.724                   \\
                            & Integrated            & 983                      & 0.698                   & \bf{1128}                     & \bf{0.747}                   & \bf{389}                      & \bf{0.920}                   & \bf{936}                      & \bf{0.765}                   \\
                            & Centralized           & 828                      & 0.851                   & 1421                     & 0.574                   & 442                      & 0.843                   & 1098                     & 0.701                   \\ \midrule
\multirow{3}{*}{Spain}      & Local                 & 298                      & 0.147                   & 37                       & 0.520                   & 41                       & -0.036                  & 231                      & 0.149                   \\
                            & Integrated            & \bf{220}                      & \bf{0.407}                   & \bf{18}                       & \bf{0.656}                   & \bf{11}                       & \bf{-0.020}                  & \bf{171}                      & \bf{0.401}                   \\
                            & Centralized           & 221                      & 0.394                   & 19                       & 0.556                   & 61                       & -0.044                  & 172                      & 0.319                   \\ \midrule
\multirow{3}{*}{Twitter}    & Local                 & 1481                     & 0.957                   & 735                      & 0.863                   & 1687                     & 0.966                   & 1299                     & 0.959                   \\
                            & Integrated            & \bf{1422}                     & \bf{0.961}                   & 735                      & 0.863                   & 1687                     & 0.966                   & \bf{1284}                     & \bf{0.961}                   \\
                            & Centralized           & 1381                     & 0.961                   & 703                      & 0.877                   & 1830                     & 0.960                   & 1330                     & 0.958                   \\ \bottomrule
\end{tabular}
\end{table}

\subsection*{ Results of traditional graph tasks}

For node classification, 
the average node classification accuracy results of 10 repeated experiments for GCN and GAT models are presented in Table \ref{tab:nodel_rel_GCN} and Table \ref{tab:nodel_rel_GAT}, respectively.

\begin{table}[!h]
\centering
\caption{Node classification performance of local, integrated, and centralized GNNs in real-world datasets incorporating CNL framework and GCN layer. The integrated models with improved performance compared to local models are highlighted in bold.}
\label{tab:nodel_rel_GCN}
\begin{tabular}{cccccc}
\toprule
Dataset                    & Model       & Agency-A & Agency-B & Agency-C & Multi-agency \\\midrule
\multirow{3}{*}{Cora}      & Local       & 0.8711   & 0.8872   & 0.9000   & 0.8723       \\
                           & Integrated  & \bf{0.8844}   & 0.8846   & \bf{0.9161}   & \bf{0.8852}       \\
                           & Centralized & 0.8723   & 0.8944   & 0.9521   & 0.8757       \\\midrule
\multirow{3}{*}{Texas}     & Local       & 0.6685   & 0.6500   & 0.4667   & 0.6546       \\
                           & Integrated  & \bf{0.7167}   & \bf{1.0000}   & \bf{0.4833}   & \bf{0.7122}       \\
                           & Centralized & 0.6995   & 0.9250   & 0.8764   & 0.7197       \\\midrule
\multirow{3}{*}{Pubmed}    & Local       & 0.9315   & 0.9039   & 0.7455   & 0.9105       \\
                           & Integrated  & \bf{0.9459}   & \bf{0.9046}   & 0.7273   & \bf{0.9146}       \\
                           & Centralized & 0.9351   & 0.8787   & 0.9181   & 0.8937       \\\midrule
\multirow{3}{*}{Wisconsin} & Local       & 0.5327   & 0.5923   & 0.6833   & 0.5646       \\
                           & Integrated  & \bf{0.5750}   & \bf{0.6846}   & \bf{0.7167}   & \bf{0.6148}       \\
                           & Centralized & 0.5873   & 0.6778   & 0.8517   & 0.6488      \\
\bottomrule
\end{tabular}
\end{table}

\begin{table}[!h]
\centering
\caption{Node classification performance of local, integrated, and centralized GNNs in real-world datasets incorporating CNL framework and GAT layer. The integrated models with improved performance compared to local models are highlighted in bold.}
\label{tab:nodel_rel_GAT}
\begin{tabular}{cccccc}
\toprule
Dataset                    & Model       & Agency-A & Agency-B & Agency-C & Multi-agency \\
\midrule
\multirow{3}{*}{Cora}      & Local       & 0.8568   & 0.8974   & 0.8355   & 0.8575       \\
                           & Integrated  & \bf{0.8659}   & \bf{0.9128}   & 0.8161   & \bf{0.8660}       \\
                           & Centralized & 0.8622   & 0.8546   & 0.9589   & 0.8655       \\\midrule
\multirow{3}{*}{Texas}     & Local       & 0.7259   & 0.8500   & 0.3333   & 0.7049       \\
                           & Integrated  & \bf{0.7389}   & \bf{1.0000}   & \bf{0.3667}   & \bf{0.7245}       \\
                           & Centralized & 0.7229   & 0.9017   & 0.7600   & 0.7279       \\\midrule
\multirow{3}{*}{PubMed}    & Local       & 0.9901   & 0.9019   & 0.8000   & 0.9236       \\
                           & Integrated  & \bf{0.9943}   & \bf{0.9032}   & \bf{0.8273}   & \bf{0.9256}       \\
                           & Centralized & 0.9487   & 0.8695   & 0.8459   & 0.8907       \\\midrule
\multirow{3}{*}{Wisconsin} & Local       & 0.5634   & 0.7000   & 0.6444   & 0.5998       \\
                           & Integrated  & 0.5500   & \bf{0.7846}   & \bf{0.6611}   & \bf{0.6085}       \\
                           & Centralized & 0.5897   & 0.7633   & 0.6267   & 0.6263      \\
\bottomrule
\end{tabular}
\end{table}

For link prediction, the GCN model is chosen as the base model, and the integrated models achieve nearly optimal performance. The average and median MAE values from 10 repeated experiments are presented in Table \ref{tab:linkResult}.

\begin{table}[!h]
\centering
\caption{Empirical prediction performance on link prediction of local, integrated, and centralized GNNs. The integrated models with improved performance compared to local models are highlighted in bold.}
\label{tab:linkResult}
\begin{tabular}{ccccc}
\toprule
Performance & Metric & Local & Integrated & Centralized \\
\midrule
 
\multirow{2}{*}{Agency-A} 
 & MAE-Mean   & \bf{0.7901} & 0.7905     & 0.7906       \\
& MAE-Median & 0.7901 & \bf{0.7900}     & 0.7906       \\
\midrule
 
\multirow{2}{*}{Agency-B} 
  & MAE-Mean   & 0.8726 & \bf{0.8712}     & 0.8737       \\
  & MAE-Median & 0.8725 & \bf{0.8712}     & 0.8738       \\
\midrule
 
\multirow{2}{*}{Agency-C} 
 & MAE-Mean   & 0.8132 & \bf{0.8119}     & 0.8132       \\
  & MAE-Median & 0.8133 & \bf{0.8122}     & 0.8132       \\
\midrule

\multirow{2}{*}{Multi-Agency} 
  & MAE-Mean   & 0.8205 & \bf{0.8198}     & 0.8210       \\
  & MAE-Median & 0.8205 & \bf{0.8197}     & 0.8210       \\

\bottomrule
\end{tabular}
\label{edge classfition}
\end{table}

\clearpage
\section*{Supplementary Note 9: Supplementary experimental settings}
\addcontentsline{toc}{section}{Supplementary Note 9: Supplementary experimental settings}

Cooperative network node service (CNNS) will automatically read the configuration file, perform necessary initialization, and open the corresponding port for the call.
It relies on Python 3.8 and includes third-party libraries such as Torch and Torch-Geometric.
In this section, we provide a detailed explanation of how hyperparameters are selected for downstream tasks.

\subsection*{Contagion dynamics}
For Twitter, influenza, and COVID-19 predictions, the batch size is set to 128, the look-back window $T$ is 20, and the horizon $h$ is 5, 10, and 14, respectively. 
We train the model using the Adam optimizer \cite{Adam} with weight decay set to 5e-4 and implement early stopping after 50 epochs if optimization does not occur to prevent overfitting.
For each task, we run 5 times with different random initializations.

\subsection*{Node classification}

For node classification experiments, we explore different learning rates ($lr$) from the set $\{0.001, 0.005, 0.01, 0.05, 0.1\}$.
We randomly partition the node set into training, validation, and testing subsets at a ratio of 60\%, 20\%, and 20\% respectively, and repeat the experiments 100 times for each dataset. 
The model optimizer used in this scenario is Adam \cite{Adam}.

\subsection*{Link prediction}

In the link prediction experiments, we set the initial learning rate to 0.003 and the weight decay to 0.001. The model optimizer employed in this scenario is Stochastic Gradient Descent (SGD). Each experiment iteration consists of 5,000 iterations, and the entire process is repeated ten times.

\end{document}